\documentclass[10pt,twocolumn,letterpaper]{article}

\usepackage{iccv}              

\definecolor{iccvblue}{rgb}{0.21,0.49,0.74}
\definecolor{green}{rgb}{0.18,0.55,0.34}
\usepackage[pagebackref,breaklinks,colorlinks,allcolors=green]{hyperref}

\usepackage{pgfplots} 

\usepackage{pgfplots} 

\usepackage{tikz}
\usepackage{xcolor}
\definecolor{blue}{RGB}{60,132,196}
\definecolor{red}{RGB}{207,78,56}
\definecolor{gray}{RGB}{146,146,161}
\definecolor{green4}{RGB}{46, 139, 87}
\definecolor{red2}{RGB}{149,9,30}
\usepackage{multirow} 
\usepackage{colortbl}
\definecolor{lightgray}{gray}{0.9}
\usepackage{hhline} 
\usepackage{pdflscape}

\newtheorem{definition}{Definition}
\usepackage{amsmath}
\usepackage{graphicx}
\usepackage{pifont}

\usepackage{dsfont}

\usepackage{svg}
\usepackage{mathrsfs}
\DeclareMathAlphabet{\mathpzc}{OT1}{pzc}{m}{it}
\definecolor{pink1}{RGB}{239 41 140}
\definecolor{pink}{RGB}{30,144,255}

\definecolor{lightpink}{RGB}{	255,20,147}

\definecolor{barriercolor}{RGB}{255, 120, 50}
\definecolor{bicyclecolor}{RGB}{255, 192, 203}
\definecolor{buscolor}{RGB}{255, 255, 0}
\definecolor{carcolor}{RGB}{0, 150, 245}
\definecolor{constructcolor}{RGB}{0, 255, 255}
\definecolor{motorcolor}{RGB}{200, 180, 0}
\definecolor{pedestriancolor}{RGB}{255, 0, 0}
\definecolor{trafficcolor}{RGB}{255, 240, 150}
\definecolor{trailercolor}{RGB}{135, 60, 0}
\definecolor{truckcolor}{RGB}{160, 32, 240}
\definecolor{driveablecolor}{RGB}{255, 0, 255}
\definecolor{otherflatcolor}{RGB}{139, 137, 137}
\definecolor{sidewalkcolor}{RGB}{75, 0, 75}
\definecolor{terraincolor}{RGB}{150, 240, 80}
\definecolor{manmadecolor}{RGB}{213, 213, 213}
\definecolor{vegetationcolor}{RGB}{0, 175, 0}
\definecolor{otherscolor}{RGB}{0, 0, 0}

\providecommand{\ie}{\textit{i.e.}}
\providecommand{\eg}{\textit{e.g.}}
\providecommand{\vs}{\textit{v.s.}}
\providecommand{\wrt}{\it{w.r.t.}}

\pgfplotsset{compat=1.18}

\usepackage{marvosym}
\usepackage{ifsym}
\def\paperID{2121} 
\def\confName{ICCV}
\def\confYear{2025}

\title{ALOcc: Adaptive Lifting-Based 3D Semantic Occupancy and \\ Cost Volume-Based Flow Predictions} 

\author{
Dubing Chen$^{1}$,
Jin Fang$^{1}$,
Wencheng Han$^{1}$,
Xinjing Cheng,\\
Junbo Yin$^{2}$,
Chenzhong Xu$^{1}$,
Fahad Shahbaz Khan$^{3,4}$,
Jianbing Shen$^{1}$\textsuperscript{\Letter}
 \\
$^1$SKL-IOTSC, CIS, University of Macau\\
$^2$CEMSE Division, King Abdullah University of Science and Technology\\
$^3$Mohamed bin Zayed University of Artificial Intelligence\quad $^4$Linköping University \\
\href{https://github.com/cdb342/ALOcc}{\textcolor{pink1}{\texttt{https://github.com/cdb342/ALOcc}}
}}

\begin{document}
\maketitle

\begin{abstract}

\let\thefootnote\relax\footnotetext{\Letter \ Corresponding author: \textit{Jianbing Shen}. This work was supported in part by the Science and Technology Development Fund of Macau SAR (FDCT) under grants 0102/2023/RIA2 and 0154/2022/A3 and
001/2024/SKL and CG2025-IOTSC, the University of Macau SRG2022-00023-IOTSC grant, and the Jiangyin Hi-tech Industrial Development Zone under the Taihu Innovation Scheme (EF2025-00003-SKL-IOTSC).}

3D semantic occupancy and flow prediction are fundamental to spatiotemporal scene understanding. This paper proposes a vision-based framework with three targeted improvements. First, we introduce an occlusion-aware adaptive lifting mechanism incorporating depth denoising. This enhances the robustness of 2D-to-3D feature transformation while mitigating reliance on depth priors. Second, we enforce 3D-2D semantic consistency via jointly optimized prototypes, using confidence- and category-aware sampling to address the long-tail classes problem. Third, to streamline joint prediction, we devise a BEV-centric cost volume to explicitly correlate semantic and flow features, supervised by a hybrid classification-regression scheme that handles diverse motion scales. Our purely convolutional architecture establishes new SOTA performance on multiple benchmarks for both semantic occupancy and joint occupancy semantic-flow prediction. We also present a family of models offering a spectrum of efficiency-performance trade-offs. Our real-time version exceeds all existing real-time methods in speed and accuracy, ensuring its practical viability.

    \end{abstract}

\section{Introduction}
    \label{sec:intro}

    \begin{figure}[t]
      \centering
      \setlength{\abovecaptionskip}{0pt}
      \includegraphics[width=0.74 \linewidth]{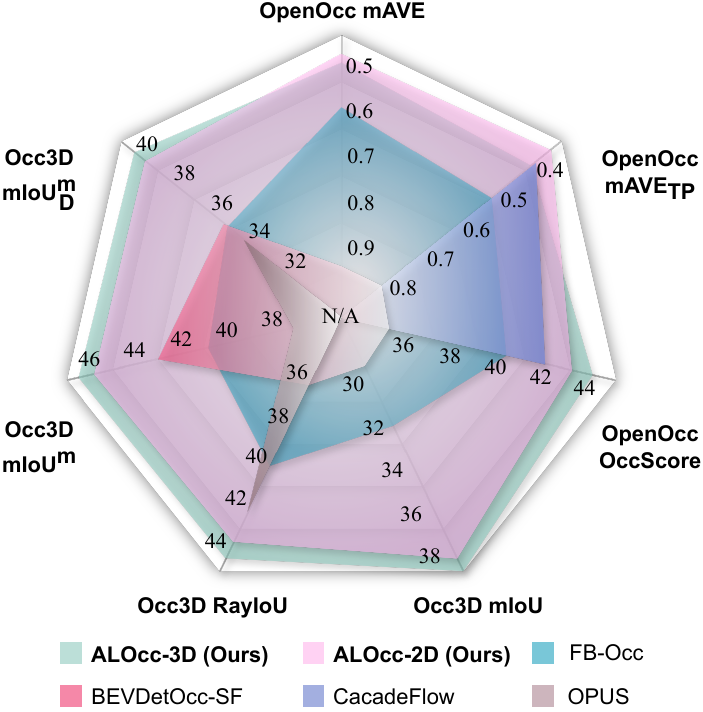}
     \vspace{1.0ex}
      \caption{\textbf{3D semantic occupancy and flow prediction results of ALOcc \vs SOTAs.} ALOcc outperforms SOTAs across various benchmarks and metrics. For a fair comparison, the input image size and backbone are standardized to $256 \times 704$ and ResNet50, with per-metric comparisons conducted under identical training conditions. Since existing methods often focus on single metrics, ``N/A" denotes unavailable values; we reimplemented FB-Occ and BEVDetOcc-SF to assess all metrics.}
      \label{fig:performance}
      \vspace{-5.2mm}
    \end{figure}
        
3D occupancy prediction is a central task in modern 3D vision, focused on converting raw camera imagery into a structured, volumetric representation of the scene~\cite{yang2017semantic,tong2023scene,huang2023tri,wang2023openoccupancy,tian2024occ3d,ma2024cam4docc,yan2025drivingsphere}. This technique produces a dense 3D grid where each voxel contains a combination of attributes, like occupancy state, semantic label, and motion flow. Such a canonical data structure for scene analysis offers significant advantages in completeness and detail over conventional, object-level descriptions like bounding boxes~\cite{li2022bevformer,li2023bevstereo,huang2022bevdet4d}.

The standard pipeline for this task, which typically involves 2D-to-3D view transformation and subsequent 3D feature decoding, faces significant challenges across its different stages. In view transformation, prevailing methods like depth-based Lift-Splat-Shoot (LSS)~\cite{philion2020lift,li2023bevstereo,huang2022bevdet4d} and cross-attention~\cite{li2022bevformer,li2023fb} present a critical trade-off. LSS is constrained by a strong depth inductive bias that can lead to premature convergence, poor handling of occlusions, and the generation of sparse features for distant objects~\cite{li2023fb}. In contrast, while cross-attention circumvents these problems, it lacks an explicit geometric foundation and performs poorly. In the decoding stage, key difficulties include robustly interpreting the reconstructed 3D volume, efficiently training on dense volumetric grids, and addressing the severe class imbalance caused by both the long-tail distribution of object classes and the predominance of \textit{empty} voxels. Furthermore, extending these frameworks to jointly predict motion flow introduces a conflicting representational demand, as features must simultaneously encode both static semantics and dynamic motion. To address these multifaceted challenges, we propose ALOcc, a unified framework that introduces targeted innovations at each stage of the occupancy prediction pipeline.

First, to overcome the limitations in view transformation, ALOcc employs an occlusion-aware adaptive lifting method. Inspired by the human cognitive ability to infer complete shapes from partial observations, this method explicitly propagates depth probabilities from visible surfaces to occluded and sparse regions, yielding a more complete and robust 3D representation. To further mitigate premature convergence from inaccurate early depth estimates, the process is stabilized by a depth denoising technique leveraging ground-truth depth data.

Next, to enhance the fidelity of the 3D decoding process, ALOcc introduces a sophisticated semantic enhancement mechanism. This mechanism correlates 3D and 2D features via a shared set of pre-defined, per-class prototypes, which transfer rich inter-class relationships from the 2D signals into the 3D representation. To address the long-tail distribution, we employ a conditional training strategy where a prototype is updated only when its class is present in the input scene. This is complemented by a hybrid sampling strategy that focuses training on hard-to-predict voxels, identified via class statistics and per-voxel prediction uncertainty. This approach significantly enhances learning efficiency.

Finally, to enable the challenging joint prediction of semantics and motion, we extend the ALOcc framework with a novel BEV cost volume-based flow head. Our core strategy here is to decouple the conflicting representational demands of this task. Specifically, we aggregate volumetric features into a compact BEV representation, from which a cost volume dedicated to motion estimation is constructed against historical features. By leveraging both this specialized cost volume and the original volumetric features for flow estimation, the primary features remain focused on occupancy semantics, thus alleviating the representational burden. A hybrid classification-regression technique is also introduced to improve adaptability to diverse motion scales.

As shown in Fig. \ref{fig:performance}, our method consistently outperforms existing approaches, establisheing new SOTA performance on multiple benchmarks for both semantic occupancy and joint occupancy semantic-flow prediction. 
Moreover, we introduce multiple variants of our ALOcc framework, employing techniques like spatial compression to achieve a superior trade-off between accuracy and computational efficiency. Our approach consistently delivers SOTA results while remaining computationally efficient, making it highly suitable for resource-constrained applications requiring real-time or on-device deployment.
In summary, our main contributions are fourfold: \textbf{\textit{(i)}} We introduce an occlusion-aware adaptive lifting method. This technique robustly projects 2D information into occluded and sparse 3D regions and is regularized by a depth denoising module to mitigate convergence to local optima. \textbf{\textit{(ii)}} We propose shared semantic prototypes that bridge the 2D and 3D domains to enhance semantic fidelity. This approach is complemented by a conditional training and uncertainty-aware sampling scheme to address severe class imbalance. \textbf{\textit{(iii)}} We introduce a novel BEV-based cost volume for occupancy flow prediction. This method alleviates the representational burden in multi-task learning and improves prediction accuracy via a hybrid classification-regression design. \textbf{\textit{(iv)}} ALOcc establishes new SOTA performance on multiple benchmarks for both semantic occupancy and joint occupancy semantic-flow prediction. This superiority extends across the efficiency spectrum, with our lightweight variant outperforming all existing real-time methods, demonstrating the framework's practical versatility.
    
\begin{figure*}[t]
    \centering
    \setlength{\abovecaptionskip}{0pt}
    \includegraphics[width=0.8\linewidth]{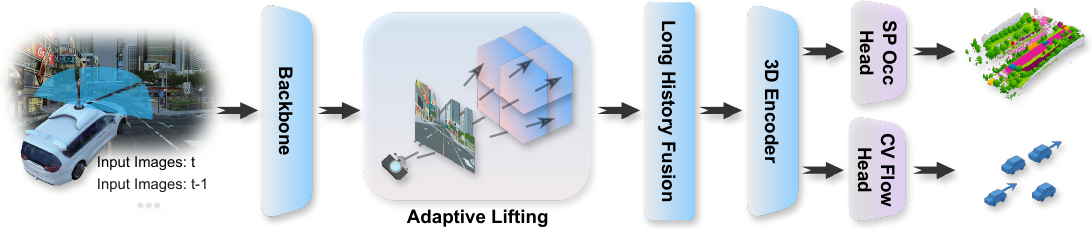}
    \vspace{1.6mm}
    \caption{\textbf{Overall Framework of ALOcc.} Image features are lifted to 3D space by adaptive lifting and encoded by a 3D encoder alongside historical features. The final representation is decoded by pluggable heads for downstream tasks (\eg, semantic occupancy, motion flow).}
    \label{fig:framework}
    \vspace{-5mm}
\end{figure*}

\section{Related Work}

\textbf{Bird's Eye View (BEV) detection}~\cite{li2022bevformer,li2022bevdepth,huang2022bevdet4d, liu2022bevfusion} has been a cornerstone for 3D scene understanding. Foundational view transformation techniques, originally developed for BEV detection, were subsequently adapted to vision-based occupancy prediction. These notably include 2D-driven methods \cite{philion2020lift, huang2022bevdet4d, li2023bevstereo}, which leverage depth maps~\cite{zheng2025decoupling} to project image features onto the BEV plane, and 3D-driven approaches \cite{li2022bevformer, li2023fb}, which employ cross-attention mechanisms to extract spatial details from 2D images.
    
 \textbf{Semantic Scene Completion (SSC)} \cite{liu2018see,li2019rgbd,chen20203d,li2023voxformer,zhang2023occformer} endeavors to reconstruct and semantically complete 3D scenes from given inputs. Early research focused on indoor scenes \cite{liu2018see,li2019rgbd,chen20203d}, predicting occupancy and semantic labels in limited scenarios. Subsequent studies gradually shifted attention to complex outdoor environments~\cite{behley2019semantickitti,cao2022monoscene}. Recent works, such as VoxFormer \cite{li2023voxformer}, adopt a two-stage approach that initially predicts occupancy, followed by semantic prediction of the occupied segments. OccFormer \cite{zhang2023occformer} introduces a dual-path Transformer submodule for 3D encoding and employs a query-based segmentation method \cite{cheng2022masked}.
    
\textbf{Vision-Based 3D Occupancy Prediction} \cite{tong2023scene,tian2024occ3d,wang2023openoccupancy,huang2023tri,chen2025rethinking} aligns closely with SSC but emphasizes multi-perspective image inputs. Early methods~\cite{huang2023tri} pioneered this task using sparse LiDAR points for supervision. Subsequently, a significant line of work \cite{tong2023scene,wei2023surroundocc,tian2024occ3d,wang2023openoccupancy} advanced the field by developing denser and more accurate annotation pipelines that leverage temporal cues and instance-level labeling. Research has focused on several key technical challenges. For the critical computational efficiency demands of dense volumetric processing, methods like OctreeOcc~\cite{lu2023octreeocc}, PanoOcc~\cite{wang2024panoocc}, SparseOcc~\cite{liu2023fully}, and OPUS~\cite{wang2024opus} have introduced techniques centered on sparse 3D representations and computations. Concurrently, to mitigate the prohibitive cost of dense 3D annotation, a line of research has explored NeRF- or 3D Gaussian-based rendering to enable supervision-light learning~\cite{pan2024renderocc,zhang2023occnerf,huang2024selfocc,liu2024letoccflow,gan2024gaussianocc}. More recently, SparseOcc~\cite{liu2023fully} explored a new metric, RayIoU, to better assess prediction quality. Furthermore, the introduction of 3D occupancy flow prediction~\cite{tong2023scene,li2024viewformer,liu2024letoccflow,liaocascadeflow,zhao20243D,chen2024adaocc} has added a temporal dimension by estimating per-voxel dynamics, pushing the boundaries of dynamic scene understanding. However, previous research lacked a cohesive evaluation framework, often conducting experiments on isolated benchmarks (such as Occ3D \cite{tian2024occ3d} or OpenOcc \cite{tong2023scene}) or comparing performance using single metrics (like mIoU or RayIoU). In this work, we present a unified framework that excels in both semantic occupancy prediction and occupancy flow prediction tasks, validated comprehensively to establish a new, robust baseline.

\section{Method}
Our research focuses on vision-based 3D semantic occupancy and flow prediction. Given $N_I$ input images $\mathbf{I} \in \mathbb{R}^{N_I \times X \times Y \times 3}$, the goal is to predict the 3D semantic occupancy grid and the 3D occupancy flow grid, supervised by the corresponding ground truth $\mathbf{O}_c \in \mathbb{R}^{H \times W \times Z}$ and $\mathbf{O}_f \in \mathbb{R}^{H \times W \times Z \times 2}$. 
Each value in $\mathbf{O}_c$ identifies the class of its corresponding spatial position, including \textit{empty} (indicating unoccupied space) and semantic classes such as \textit{pedestrian} and \textit{car}. 
Each voxel in the flow grid $\mathbf{O}_f$ contains a 2D flow vector specifying movement in the $x$ and $y$ directions.
\cref{fig:framework} presents a schematic view of our approach.
Initially, 2D features $\mathbf{f}_I$ are extracted from surrounding images, which are subsequently transformed into 3D space by adaptive lifting. The lifted features $\mathbf{f}_{Lift}$ are encoded within the 3D space, together with historical frame data. 
The encoded volume features $\mathbf{f}_{v}$ are then decoded by various task heads to predict the 3D semantic occupancy and flow grids. 

Overall, Sec. \ref{sec:ada-lift} presents an occlusion-aware adaptive lifting technique to improve 2D-to-3D view transformation.
Sec. \ref{sec:proto} describes a semantic prototype-based occupancy prediction head that enhances 2D-3D semantic alignment and addresses the long-tail problem. 
Sec. \ref{sec:flow} introduces a BEV cost volume approach with a regression-classification strategy to boost occupancy flow predictions.

\subsection{Revisiting Depth-Based LSS}
\label{sec:lss}
As the core module of vision-based occupancy prediction, the 2D-to-3D view transformation process is formulated as:
\begin{align}
    \mathbf{f}_{Lift} = \mathbf{M}_T \cdot   \mathbf{f}_I ,
    \label{eq:2dto3d}
    \end{align}
where $\mathbf{f}_I$ and $\mathbf{f}_{Lift}$ denote the image and lifted features, respectively. Flattening and reshaping operations are omitted here.
Depth-based LSS initializes $\mathbf{M}_T$ as a zero matrix, filling it with discretized depth probabilities. For each pixel feature $\mathbf{x}^{(u,v)}$, depth probability is predicted as $\{P(o^i_d|\mathbf{x})\}_{i=1}^{D}$ across $D$ discretized bins along the camera sightline. 
Assume bin \(i\)'s true depth is \(d\) (with slight abuse of notation). 
The coordinate is mapped with camera parameters (provided by datasets) from the image system (ICS) to the voxel system (VCS), \ie, \((u,v,d) \to (h,w,z)\). Each depth probability is assigned to \(\mathbf{M}_T\) at the row indexed by the rounded voxel coordinate $[(h, w, z)]$ and the column indexed by $(u,v)$. Given $\mathbf{M}_T$’s sparsity, this is efficiently implemented using sparse operations, such as \textit{sparse\_coo\_tensor} in \textit{PyTorch} or BEVPool \cite{philion2020lift, huang2022bevdet4d}.

\subsection{Occlusion-Aware Adaptive Lifting}
\label{sec:ada-lift}
\begin{figure}[t]
    \centering
    \setlength{\abovecaptionskip}{0pt}
    \includegraphics[width=0.98\linewidth]{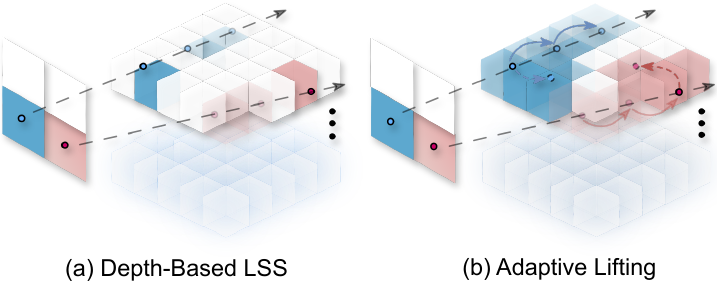}
    \vspace{0.9mm}
    \caption{\textbf{Comparison between \textbf{(a)} Depth-based LSS and \textbf{(b)} Our occlusion-aware adaptive lifting.} Blue and red denote distinct pixels or voxels. The dotted line represents the camera sightline, with opacity reflecting the weight of mapping 2D features to 3D voxels—higher opacity indicates greater weights.
    }
    \label{fig:ada-lift}
    \vspace{-4.4mm}
\end{figure}

We enhance depth-based LSS by introducing probability transfer from surface to occluded areas. We first replace the hard rounding-based filling strategy in Sec.~\ref{sec:lss} with a probability-based soft filling approach. As detailed in Fig.~\ref{fig:ada-lift}, we use trilinear interpolation to diffuse a mapped coordinate in VCS over its eight neighboring points. Continuing the notation from the previous section, the interpolation calculates eight probability values:
\begin{align}
\resizebox{.905\linewidth}{!}{
$\{p_{s} = (1 - |h - \hat{h}_s|) \cdot (1 - |w - \hat{w}_s|) \cdot (1 - |z - \hat{z}_s|)\}_{s=1}^8$,
}
\label{eq:p_i}
\end{align}
where $\{(\hat{h}_s, \hat{w}_s, \hat{z}_s)\}_{s=1}^8$ are neighbor coordinates. For each of them, we fill \(\mathbf{M}_T\) with \(p_s \cdot P(o^i_d|\mathbf{x})\), indexing the column with the ICS coordinate $(u,v)$ and the row with VCS coordinate \((\hat{h}_s, \hat{w}_s, \hat{z}_s)\). This distributes \(P(o^i_d|\mathbf{x})\) by distance, enabling differentiable 2D-to-3D lifting \textit{\wrt} coordinates.

Depth-based LSS methods excel by guiding the 2D-to-3D feature transformation through explicit depth probability modeling.
However, the target of depth estimation follows a $\delta$ distribution, which causes most of the weights to concentrate on surface points. As a result, occluded areas get much lower weights, creating a bias that hinders learning from these regions. We aim to resolve this informational gap by addressing two key scenarios: intra-object (within an object) and inter-object (between objects) occlusion. 
For both, we construct a probability transfer matrix from visible to occluded parts. 
For the former type of occlusion, we design conditional probabilities to transfer surface probabilities to occluded length probabilities. 
For notational simplicity, we redefine the discrete depth probabilities of pixel feature $\mathbf{x}$ as $\{P(o^i_d)\}_{i=1}^D$. We employ the same binning approach to discretize occluded length predictions. To convert the discrete depth probability $P(o_d)$ into the discrete occluded length probability $P(o_{ol})$, we introduce the Bayesian conditional probability $P(o_{ol} | o_d)$:
\begin{align}
\centering
P(o^j_{ol}) &= \sum_{i=1}^{D} P(o^j_{ol} | o^i_d) \cdot P(o_d^i), \quad j = 1, \ldots, D .
\label{eq:p_o}
\end{align}

To determine $P(o^j_{ol} | o_d^i)$, we consider the physical meaning of depth: if the camera sightline reaches point $i$, it indicates that all points at shallower depths (\ie, with indices smaller than $i$ are empty.
Therefore, we only need to parameterically estimate the conditional probability for bin $i$ to positions with larger depth. $P(o^j_{ol} | o_d^i)$ is modeled as:
\begin{align}
\centering
P(o^j_{ol} | o_d^i) &= 
\begin{cases} 
0, & \text{if } j < i \\
1, & \text{if } j=i \\
f_{ol}(\mathbf{x}, j - i), & \text{if } j > i .
\end{cases}
\label{eq:p_o_d}
\end{align}
Here, $f_{ol}(\mathbf{x}, j - i)$ is the likelihood of bin $j$ being occupied given bin $i$ corresponds to the surface point, estimated using pixel feature $\mathbf{x}$. In implementation, we compute $(D-1)$ discrete probabilities across all relative positions of $i$ and $j$ via convolution. The transformation from depth to occluded length probabilities is achieved by multiplying depth probabilities with the causal conditional probability matrix.

For inter-object occlusion, we design a probability transfer matrix to propagate depth probabilities to surrounding points. For each point $(u, v, d)$, we use MLPs to estimate offsets $(\Delta u, \Delta v)$ and weights $\omega$ from its feature $\mathbf{x}$. New points $(u + \Delta u, v + \Delta v, d^i; \omega \cdot P(o_{ol}))$ are then incorporated into $\mathbf{M}_T$ via soft filling and optimized during training. For efficiency, we limit transfer to the $m$ points with top depth probabilities. $m$ is set to a small value of 3 in practice.

The predominant role of depth probabilities in 2D-to-3D transformation can lead to suboptimal model convergence due to initial inaccurate depth estimations. To mitigate this, we introduce a denoising operation, inspired by query denoising in object detection \cite{li2022dn,zhang2022dino,he2024real}. 
This approach utilizes ground-truth depth probabilities to guide early training. 
We combine ground-truth and predicted depths via a weighted average for adaptive lifting, with the ground-truth weight starting at 1 and decaying to 0 via cosine annealing:
\begin{equation}
\resizebox{.905\linewidth}{!}{$
P(o_d)\!=\!\frac{1}{2} [(1\!+\!\cos(\frac{\pi e}{E})\!\cdot\!P_{gt}(o_d)\!+\!(1 \!-\! \cos(\frac{\pi e}{E}))\!\cdot\!P_{pred}(o_d)] ,
$}
\end{equation}
where $P(o_d)$ is the final depth probability for adaptive lifting. $P_{gt}(o_d)$ and $P_{pred}(o_d)$ are ground truth and predicted depth probabilities, $e$ is the current training step, and $E$ is the total annealing steps. During test, $P(o_d) =P_{pred}(o_d)$.


\subsection{Semantic Prototype-Based Occupancy Head}
\label{sec:proto}
\begin{figure}[t]
    \centering
    \setlength{\abovecaptionskip}{0pt}
    \includegraphics[width=1\linewidth]{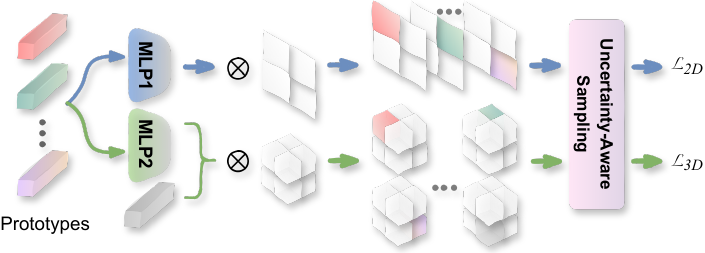}
    \vspace{-2.5mm}
    \caption{\textbf{Illustration of the semantic prototype-based occupancy head.} Shared semantic prototypes calculate segmentation logits for both 2D and 3D features. Training is conducted only on the top $K$ voxels (or pixels) sampled based on logit uncertainty.}
    \label{fig:occ-head}
    \vspace{-4.4mm}
\end{figure}

After 2D-to-3D view transformation, we align features across domains using shared semantic prototypes (Fig.~\ref{fig:occ-head}). These prototypes are randomly initialized and function as the class weights for both 2D and 3D loss computations, thereby creating a direct semantic bridge between the 2D and 3D feature representations.
Given per-class prototypes, a straightforward method for decoding semantic occupancy is to compute voxel-feature-to-prototype similarity with cross-entropy supervision. However, this is suboptimal due to the severe class imbalance in typical scenes. 

To overcome this, we introduce a targeted training strategy comprising two components: conditional loss computation and uncertainty-guided voxel sampling. First, we only compute losses for classes present in each ground-truth sample, generating per-class logit maps by computing the inner product between the voxel features against the prototypes of only these classes. Second, inspired by prior works~\cite{cheng2022masked,zhang2023occformer,kirillov2020pointrend}, we perform uncertainty-guided sampling to further focus the training on informative regions. We treat the computed logits as a measure of model uncertainty and combine this with class priors to form a sampling distribution. From this distribution, we sample $K$ hard voxels, concentrating computation on low-confidence regions and under-represented classes. The final loss is then computed exclusively on the sampled voxels and is a combination of per-class binary cross-entropy and Dice loss:
\begin{align}
\mathpzc{L}_{3D} = \alpha \mathpzc{L}_{Dice} + \beta \mathpzc{L}_{BCE},
\end{align}
where $\alpha, \beta$ are balancing coefficients. During inference, the final voxel category prediction is the one with the highest response to its semantic prototype or \textit{empty} embedding:
\begin{align}
\hat{\mathbf{o}}_v = \textrm{arg}\max_c (\textrm{MLP}({P_c}) \cdot \mathbf{f}_v ),
\end{align}
where $\mathbf{f}_v$ denotes the volume features, $P_c$ is the prototype of class $c$ or \textit{empty} embedding, and $\textrm{MLP}$ further encodes $P_c$.
To further enforce semantic consistency, we also introduce an auxiliary 2D loss, $\mathpzc{L}_{2D}$, by supervising the 2D image features with the same shared prototypes. The 2D ground-truth masks for this loss are generated by projecting semantic LiDAR data (a component of the occupancy GT) onto the image planes. The formulation of $\mathpzc{L}_{2D}$ mirrors that of $\mathpzc{L}_{3D}$, combining Dice and binary cross-entropy losses.

\subsection{BEV Cost Volume-Based Flow Prediction}
\label{sec:flow}
\begin{figure}[t]
    \centering
    \setlength{\abovecaptionskip}{0pt}
    \includegraphics[width=1\linewidth]{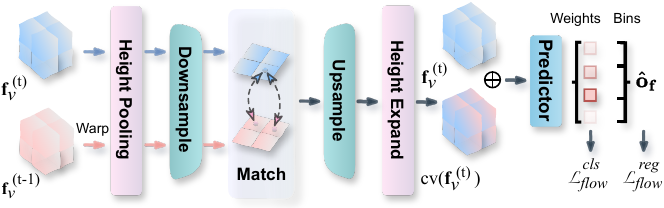}
    \vspace{-2mm}
    \caption{\textbf{Illustration of the Flow Head.} The cost volume is generated from downsampled BEV features and used to predict bin probabilities along with volume features. Final flow values are computed as a weighted sum of the bin center values.}
    \label{fig:flow-head}
    \vspace{-4.4mm}
\end{figure}

As shown in \cref{fig:framework}, occupancy flow is decoded from the same volumetric features as semantics. Conventional approaches typically predict flow using single-frame features. However, this places a significant representational bottleneck, as features must joinly encode static semantics and dynamic motion cues. Inspired by stereo matching~\cite{li2023bevstereo}, we resolve this by constructing a BEV cost volume, which provides an explicit motion prior from cross-frame correspondences to decouple the tasks. 
The cost volume construction (\cref{fig:flow-head}) begins by collapsing the volumetric features (0-4m height) into a compact, downsampled BEV map. This step is crucial as it efficiently models the dominant planar (x-y) motion while simultaneously enlarging the features' receptive fields. The prior frame's BEV map is then warped to the current coordinate system via ego-motion. Finally, the cost volume is populated by computing the cosine similarity between current features and warped historical features within a local search window~\cite{li2023bevstereo}:
\begin{align} 
    \mathrm{cv}(\mathbf{f}_{v}^{(t)}; k) = \frac{\hat{\mathbf{f}}_{v}^{(t)} \cdot \mathrm{warp}(\mathbf{\hat{f}}_{v}^{(t-1)}(\Delta p_k))}{\|\hat{\mathbf{f}}_{v}^{(t)}\|_2 \cdot \|\mathrm{warp}(\hat{\mathbf{f}}_{v}^{(t-1)}(\Delta p_k))\|_2}, 
\end{align}
where $\hat{\mathbf{f}}_{v}^{(t)}$ is the current frame’s BEV map, derived by compressing and downsampling the volumetric feature $\mathbf{f}_{v}^{(t)}$. The $\mathrm{warp}(\cdot)$ function transforms historical features to the current coordinate system, and ${\Delta p_k}$ denotes a predefined sampling offset in the search window. For notational simplicity, we omit post-processing steps like upsampling from this formulation; these operations are illustrated in \cref{fig:flow-head}. The final flow prediction head is then formulated as:
\begin{align}
\hat{\mathbf{o}}\mathbf{_f} = f_{\mathit{Flow}}(\mathrm{cat}(\mathbf{f}_{v}^{(t)}, \mathrm{cv}(\mathbf{f}_{v}^{(t)}))).
\end{align}

The cost volume-based flow prediction method establishes cross-frame correspondences through explicit feature matching. This approach alleviates the representational burden inherent in semantic-motion joint prediction and is highly efficient, as it reuses cached features from the prior frame, obviating the need for re-computation.
Furthermore, to enhance robustness to diverse flow magnitudes, we adopt a hybrid classification-regression scheme \cite{bhat2021adabins,li2024binsformer}. 
The continuous flow space is discretized into a set of bins based on the value range observed in the training data. The flow head then predicts a probability distribution over these bins, from which the final continuous flow vector is computed as the expectation over the bin centers:
\begin{align}
\hat{\mathbf{o}}\mathbf{_f} = \sum_{n=1}^{N_b} p_b^n \cdot \mathbf{b}^n ,
\end{align}
where $p_b^n$ is the predicted probability of the flow vector falling into the $n$-th bin, $\mathbf{b}^n$ is the corresponding bin's center value over a total of $N_b$ bins. $f_{\mathit{Flow}}$ is thus redefined to predict $p_b$. We supervise the prediction with the ground-truth flow $\mathbf{o_f}$ by minimizing the L2 loss and maximizing the cosine similarity:
\begin{align}
\mathpzc{L}_{flow}^{reg} = \| \hat{\mathbf{o}}\mathbf{_f} - \mathbf{o_f} \|_2^2 -\frac{\hat{\mathbf{o}}_\mathbf{f} \cdot \mathbf{o_f}}{\| \hat{\mathbf{o}}_\mathbf{f} \|_2 \cdot \| \mathbf{o_f} \|_2} ,
\label{eq:flow_loss_reg}
\end{align}
where the L2 loss ensures magnitude accuracy while cosine similarity guarantees directional precision. To improve flow classification across discrete bins, we additionally introduce a classification loss with the ground-truth bin index:
\begin{align}
    \mathpzc{L}_{flow}^{cls} = -\sum_{n=1}^{N_b}\mathds{1}(\text{Index}[\mathbf{o_f};\mathbf{b}]=n) \log(p_b^n) ,
\end{align}
where $\mathds{1}(\cdot)$ is the indicator function and $\text{Index}[\mathbf{o_f};\mathbf{b}]$ represents the index of ground truth $\mathbf{o_f}$ within the bins $\mathbf{b}$.

The final occupancy flow loss is:
\begin{align}
\mathpzc{L}_{flow} = \mathpzc{L}_{flow}^{reg} +  \mathpzc{L}_{flow}^{cls} .
\end{align}

\begin{figure}[t]
    \centering
    \setlength{\abovecaptionskip}{0pt}
    \includegraphics[width=0.8\linewidth]{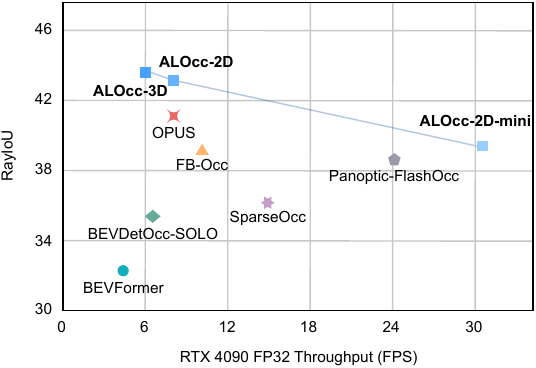}
    \caption{\textbf{Speed/accuracy trade-off of our method.} We benchmark RayIoU and speed on the Occ3D dataset using one RTX 4090 GPU with a batch size of 1. All the methods utilize a ResNet-50 backbone and 256$\times$704 input size, except BEVFormer (900$\times$1600). All detailed results are available in \cref{tab:sota-womask}. }
    \label{fig:trade-off}
    \vspace{-5mm}
\end{figure}

\subsection{Training Objective}
For only predicting 3D semantic occupancy, the overall training objective is
\vspace{-2mm}
\begin{align}
\mathpzc{L}_{sem} = \mathpzc{L}_{3D} +\mathpzc{L}_{2D}+\mathpzc{L}_{depth} ,
\end{align}
where $\mathpzc{L}_{depth}$ is the depth loss. When extending to joint semantic occupancy and flow prediction, we incorporate flow supervision, resulting in the overall objective:
\vspace{-2mm}
\begin{align}
\mathpzc{L}_{sem-flow} = \mathpzc{L}_{3D} +\mathpzc{L}_{2D}+ \mathpzc{L}_{depth}+\mathpzc{L}_{flow} .
\end{align}
\vspace{-2mm}

\begin{table}[t]
  
      \setlength{\abovecaptionskip}{0pt}
      \setlength{\tabcolsep}{0.015\linewidth}
      \begin{center}
      \resizebox{0.48\textwidth}{!}{
      \begin{tabular}{l|c|c|c|c|c|c}
          \toprule
           Method  &Input& Backbone & Image Size  & mIoU\textsubscript{D}\textsuperscript{m} & mIoU\textsuperscript{m}  & FPS\\
          \midrule
  
           BEVDetOcc-SF~\cite{huang2022bevdet4d}  &C& ResNet-50 & $256\times704$ &34.4& 41.9 &6.5 \\
             
             UniOCC~\cite{pan2023uniocc}  &C& ResNet-50 & $256\times704$  &- & 39.7 &-\\
             FB-Occ~\cite{li2023fbocc}  &C& ResNet-50 & $256\times704$  &34.2 & 39.8 &10.3 \\
            
            SurroundSDF~\cite{liu2024surroundsdf} &C & ResNet-50 & $256\times704$ &36.2 & 42.4 &-\\
            FlashOCC~\cite{yu2023flashocc}  &C& ResNet-50 & $256\times704$ &24.7& 32.0 &29.6  \\
            COTR~\cite{ma2023cotr} &C & ResNet-50 & $256\times704$  &\underline{38.6} & \underline{44.5} &0.5\\
              
              ViewFormer~\cite{li2024viewformer}&C& ResNet-50 & $256\times704$   &35.0 & 41.9 &-\\
              OPUS~\cite{wang2024opus}  &C& ResNet-50 & $256\times704$ &33.3 &36.2 &8.2 \\
            \rowcolor{pink!10} \textbf{ALOcc-2D-mini}  &C& ResNet-50 & $256\times704$ &35.4& 41.4  &30.5\\
            \rowcolor{pink!10} \textbf{ALOcc-2D}  &C& ResNet-50 & $256\times704$ &38.7{\color{red2}~$\uparrow$0.1}& 44.8{\color{red2}~$\uparrow$0.3}  &8.2\\
           \rowcolor{pink!10} \textbf{ALOcc-3D}   &C& ResNet-50 & $256\times704$ &\textbf{39.3}{\color{red2}~$\uparrow$0.7} &\textbf{45.5}{\color{red2}~$\uparrow$1.0} &6.0 \\
          \rowcolor{lightpink!8} \textbf{ALOcc-2D-mini}  &C& Intern-T & $256\times704$ &37.9& 43.7  &16.6\\
            \rowcolor{lightpink!8} \textbf{ALOcc-2D}  &C& Intern-T & $256\times704$ &40.7& 46.6 &8.8\\
           \rowcolor{lightpink!8} \textbf{ALOcc-3D}   &C& Intern-T & $256\times704$ &41.5&47.5 &5.9\\
           
           \midrule
           HyDRa~\cite{wolters2024unleashing} &C+R& ResNet-50 & $256\times704$  &40.6 & 44.4 &-\\
           EFFOcc~\cite{shi2024effocc} &C+L& ResNet-50 & $256\times704$   &\underline{50.1} & \underline{52.8} &-\\
           SDGOcc~\cite{duan2025sdgocc} &C+L& ResNet-50 & $256\times704$   &47.7 & 51.7 &7.5\\

           \rowcolor{pink!10} \textbf{ALOcc-2D-mini}  &C+D& ResNet-50 & $256\times704$ &46.2& 50.0 &28.1\\
            \rowcolor{pink!10} \textbf{ALOcc-2D}  &C+D& ResNet-50 & $256\times704$ &50.3{\color{red2}~$\uparrow$0.2}& 53.5{\color{red2}~$\uparrow$0.7}  &8.1\\
           \rowcolor{pink!10} \textbf{ALOcc-3D}   &C+D& ResNet-50 & $256\times704$ &\textbf{50.6}{\color{red2}~$\uparrow$0.5} &\textbf{54.5}{\color{red2}~$\uparrow$1.7} &6.0 \\
          \rowcolor{lightpink!8} \textbf{ALOcc-2D-mini}  &C+D& Intern-T & $256\times704$ &48.9& 52.1  &16.1\\
            \rowcolor{lightpink!8} \textbf{ALOcc-2D}  &C+D& Intern-T & $256\times704$ &52.0& 54.9 &8.7\\
           \rowcolor{lightpink!8} \textbf{ALOcc-3D}   &C+D& Intern-T & $256\times704$ &52.4&55.6 &5.8\\
         
           \midrule
           \midrule
           BEVFormer~\cite{li2022bevformer} &C & ResNet-101 & $900\times1600$ &37.2 & 39.2  &4.4\\
           
           VoxFormer~\cite{li2023voxformer}  &C& ResNet-101 & $900\times1600$  &-& 40.7  &-\\
           SurroundOcc~\cite{wei2023surroundocc} &C & ResNet-50 & $900\times1600$ &31.2 & 37.2 &-\\
    FastOcc~\cite{hou2024fastocc}  &C& ResNet-101 & $640\times1600$ &34.5 &39.2  &-\\
            PanoOcc~\cite{wang2024panoocc}  &C& ResNet-101 & $640\times1600$ &37.3 & 42.1  &-\\
            OSP~\cite{shi2024occupancy}&C& ResNet-101 & $900\times1600$  &37.0 & 41.2 &-\\
           BEVDetOcc~\cite{huang2022bevdet4d}  &C& Swin-Base & $512\times1408$ &36.9& 42.0 &1.1 \\
  
           COTR~\cite{ma2023cotr}  &C& Swin-Base & $512\times1408$ &\underline{41.3} &\underline{46.2}  &-\\

           \rowcolor{pink!10} \textbf{ALOcc-2D} &C&Swin-Base & $512\times1408$  &44.5{\color{red2}~$\uparrow$3.2} &49.3{\color{red2}~$\uparrow$3.1} & 1.8 \\
           \rowcolor{pink!10} \textbf{ALOcc-3D} &C &Swin-Base & $512\times1408$   &\textbf{46.1}{\color{red2}~$\uparrow$4.8} & \textbf{50.6}{\color{red2}~$\uparrow$4.4} & 1.5 \\

           \midrule
           OccFusion~\cite{ming2024occfusion} &C+L& ResNet-101 & $900\times1600$   & 45.3  &46.8 &-\\
           BEVFusion~\cite{liu2022bevfusion} &C+L& Swin-Base & $512\times1408$  &48.7 & 54.0 &-\\
           EFFOcc~\cite{shi2024effocc} &C+L& Swin-Base & $512\times1408$  &50.7 & 54.1 &-\\
FusionOcc~\cite{zhang2024fusionocc} &C+L& Swin-Base & $512\times1408$  &\underline{53.1} & \underline{56.6} &-\\
\rowcolor{pink!10} \textbf{ALOcc-2D} &C+D&Swin-Base & $512\times1408$  &56.8{\color{red2}~$\uparrow$3.7} &58.7{\color{red2}~$\uparrow$2.1} & 1.8\\
           \rowcolor{pink!10} \textbf{ALOcc-3D} &C+D &Swin-Base & $512\times1408$   &\textbf{57.8}{\color{red2}~$\uparrow$4.7} & \textbf{60.0}{\color{red2}~$\uparrow$3.4} & 1.5 \\
           
          \bottomrule
      \end{tabular}
      }
      \end{center}
      \vspace{-2mm}
      \caption{\textbf{3D semantic occupancy prediction performance of mIoU\textsubscript{D}\textsuperscript{m} and mIoU\textsuperscript{m} \textit{\wrt} training \textit{with} camera visible mask on Occ3D.} Input modalities include \textit{Camera (C)}, \textit{Radar (R)}, \textit{LIDAR (L)}, and \textit{Depth (D)}, a sparser signal mapped from LIDAR points. Best results among similar conditions are \textbf{bolded}, top results from other methods are \underline{underlined}, with improvements over these marked by arrows. FPS is measured on RTX 4090.
      }
      \label{tab:sota}
        \vspace{-5mm}
  \end{table}

\vspace{-2.4mm}
\section{Experiment}
\subsection{Experimental Setup}
\textbf{Dataset.} 
We employ the nuScenes dataset \cite{caesar2020nuScenes} for most experiments, which contains 700 training, 150 validation, and 150 test scenes. Occ3D \cite{tian2024occ3d} and OpenOcc \cite{tong2023scene} add voxel-wise annotations to nuScenes, spanning a range of $\pm40m$ (X,Y) and $-1m$ to $5.4m$ (Z). 
The voxel resolution is $0.4m$ in all dimensions. 
Occ3D provides 18 semantic categories, including 17 object classes and one \textit{empty} class that denotes unoccupied spaces. 
OpenOcc provides 17 categories (omitting ``others") and additionally includes per-voxel motion flow annotations for the X-Y plane.

\begin{table*}[t]
  \setlength{\tabcolsep}{5pt}
   \centering
  \setlength{\tabcolsep}{2.8mm}{
   \scalebox{0.78}{
   \begin{tabular}{l|c|c|c|c|ccc|c}
      \toprule
      Method &Backbone & Image Size & mIoU & {RayIoU} & \multicolumn{3}{c|}{RayIoU\textsubscript{1m, 2m, 4m}}  & FPS \\
      \midrule
     
      OccFormer~\cite{zhang2023occformer}  & ResNet-101 & $640\times960$  & 21.9& -& -& -& -& -   \\
      TPVFormer~\cite{huang2023tri}  & ResNet-101 & $640\times960$  & 27.8& -& -& -& - & -  \\
      CTF-Occ~\cite{tian2024occ3d}  & ResNet-101 & $640\times960$  & 28.5& - & -& -& -& -  \\
       BEVFormer~\cite{li2022bevformer}  & ResNet-101   & 900$\times$1600 & 23.7& 32.4 & 26.1 & 32.9 & 38.0  & 4.4 \\
      
      BEVDetOcc-SF \cite{huang2022bevdet4d}    & ResNet-50    & 256$\times$704  & 24.3 & 35.2 & 31.2 & 35.9 & 38.4  &6.5 \\
      FB-Occ \cite{li2023fbocc}      & ResNet-50    & 256$\times$704  & 31.1& 39.0 & 33.0 & 40.0 & 44.0  & 10.3\\
     
     RenderOcc \cite{pan2024renderocc} & Swin-Base & 512$\times$1408  &24.4  & 19.5 & 13.4 & 19.6 & 25.5 & - \\
      SparseOcc \cite{liu2023fully}          & ResNet-50    & 256$\times$704   & 30.9 & {36.1} & 30.2 & 36.8 &41.2  & 15.0 \\
      Panoptic-FlashOcc \cite{yu2024panoptic}          & ResNet-50    & 256$\times$704   & \underline{31.6} & 38.5 &32.8 &39.3 &43.4  & 24.2 \\
      OPUS \cite{wang2024opus}          & ResNet-50    & 256$\times$704   & - & \underline{41.2} & 34.7 &42.1 &46.7  & 8.2 \\

      \rowcolor{pink!10} \textbf{ALOcc-2D-mini}   & ResNet-50    & 256$\times$704 & 33.4 & 39.3 & 32.9 &40.1 & 44.8  &30.5  \\
      \rowcolor{pink!10} \textbf{ALOcc-2D}   & ResNet-50    & 256$\times$704 & 37.4{\color{red2}~$\uparrow$5.8} & 43.0{\color{red2}~$\uparrow$1.8} & 37.1 &43.8 & 48.2  &8.2  \\

      \rowcolor{pink!10} \textbf{ALOcc-3D}   & ResNet-50    & 256$\times$704   & \textbf{38.0}{\color{red2}~$\uparrow$6.4} & {\textbf{43.7}{\color{red2}~$\uparrow$2.5}} & \textbf{37.8} & \textbf{44.7} & \textbf{48.8}  &6.0  \\
      \rowcolor{lightpink!8} \textbf{ALOcc-2D-mini}   & Intern-T    & 256$\times$704 & 35.9 & 42.4 &  35.9&43.3& 47.9 & 16.6 \\
      \rowcolor{lightpink!8} \textbf{ALOcc-2D}   & Intern-T    & 256$\times$704 & 39.1 & 44.9 & 38.8 &45.8 & 50.0  &8.8  \\

      \rowcolor{lightpink!8} \textbf{ALOcc-3D}   & Intern-T    & 256$\times$704   & 40.0& 45.9 & 39.8 & 46.9 & 50.9 &5.9  \\
      
      \bottomrule
   \end{tabular}
   }
   }
   \vspace{-2mm}
    \caption{\textbf{3D semantic occupancy prediction performance of RayIoU and mIoU \textit{\wrt} training \textit{without} camera visible mask on Occ3D.} All methods operate on camera inputs.}
   \label{tab:sota-womask}
   \vspace{-2.4mm}
\end{table*}

\begin{table*}[t]
  \setlength{\tabcolsep}{5pt}
   \centering
   \setlength{\tabcolsep}{2.2mm}{
   \scalebox{0.80}{
   \begin{tabular}{l|c|c|c|c|c|c|c|ccc|c}
      \toprule
      Method &Sup. &Backbone & Image Size & Occ Score &mAVE &mAVE\textsubscript{TP} & {RayIoU} & \multicolumn{3}{c}{RayIoU\textsubscript{1m, 2m, 4m}} &FPS \\
      \midrule
     
      OccNeRF~\cite{zhang2023occnerf}  &2D& ResNet-101 & 256$\times$704  & 28.5 &-&  1.59  & 31.7   &16.6 &29.3 &49.2   &-\\
      RenderOcc~\cite{pan2024renderocc}  &2D&Swin-Base & 512$\times$1408  & 33.0 &-& 1.63  & 36.7    &20.3 &32.7 &49.9   &- \\
      
      LetOccFlow~\cite{liu2025let} &2D & ConvNeXt-Base & 512$\times$1408  & 36.4 &-&1.45  & \underline{40.5}     &25.5 &39.7 &56.3  &-  \\
      OccNet~\cite{tong2023scene}  &3D& ResNet-50 & 900$\times$1600  & 35.7&-& 1.61  & 39.66   &29.3 &39.7 &50.0  &- \\
      BEVDetOcc-SF~\cite{huang2022bevdet4d} &3D & ResNet-50 & 250$\times$704 &33.0 &1.42 &1.78 &36.7  &31.6 &37.3 &41.1 &6.2\\
      FB-Occ~\cite{li2023fbocc} &3D & ResNet-50 & 250$\times$704 &39.2 &\underline{0.591} &0.651 &39.0  &32.7 &39.9 &44.4 &10.1\\
      CascadeFlow~\cite{liaocascadeflow} &3D & ResNet-50 & 250$\times$704  & 40.9&-& \underline{0.47} &39.6  &33.5 &40.6 &45.3 &-   \\
    F-Occ~\cite{zhao20243D}&3D & Intern-T  & 250$\times$704  &\underline{41.0} &- &0.493 &39.9 &33.9 &40.7 &45.2 &- \\

       \rowcolor{pink!10} \textbf{ALOcc-Flow-2D}&3D & ResNet-50    & 256$\times$704    & 42.1{\color{red2}~$\uparrow$1.1}& {\textbf{0.537}}{\color{red2}~$\downarrow$0.054} & \textbf{0.427}{\color{red2}~$\downarrow$0.043} &40.5&  34.3 & 41.3& 45.8 & 7.0\\

      \rowcolor{pink!10} \textbf{ALOcc-Flow-3D} &3D  & ResNet-50    & 256$\times$704    & \textbf{43.0}{\color{red2}~$\uparrow$2.0}& 0.556{\color{red2}~$\downarrow$0.035} & 0.481 &\textbf{41.9}{\color{red2}~$\uparrow$1.4}&\textbf{35.6}&\textbf{42.8} & \textbf{47.4}  & 5.5 \\

      \rowcolor{lightpink!8} \textbf{ALOcc-Flow-2D} &3D& Intern-T    & 256$\times$704    & 44.4& 0.545 & 0.451 &43.2&  36.8 & 44.1& 48.7& 6.7\\

      \rowcolor{lightpink!8} \textbf{ALOcc-Flow-3D}  &3D & Intern-T    & 256$\times$704    & 45.3& 0.574 & 0.469 &44.4&38.0&45.5 & 49.8 &5.3  \\
      
      \bottomrule
   \end{tabular}
   }
   }
   \vspace{-2mm}
   \caption{\textbf{Performance evaluation of joint 3D semantic occupancy and flow prediction on OpenOcc.} All methods listed are camera-based. The ``sup." column indicates the supervision signal, specifying that weakly supervised methods are trained using only labels.}

   \label{tab:sem-flow}
   \vspace{-5mm}
\end{table*}

\noindent\textbf{Benchmarks.} 
We evaluate ALOcc on three distinct benchmarks to assess its capabilities in various scenarios: \textit{i) 3D semantic occupancy prediction with mask.} 
This benchmark emphasizes the prediction of observed regions. 
Following prior works~\cite{wang2024panoocc,ma2023cotr,li2023fbocc}, ALOcc is trained on Occ3D using the official camera visibility mask. Performance is measured by mIoU on 17 semantic categories. 
To better reflect performance on critical dynamic objects, we also report mIoU\textsubscript{D} on a subset of 8 dynamic foreground classes. \textit{ii) 3D semantic occupancy prediction without mask.} This benchmark emphasizes the simultaneous prediction of both observed and occluded regions. We follow the setup of SparseOcc~\cite{liu2023fully} and train on Occ3D without the visibility masks. We calculate mIoU and RayIoU for 17 semantic categories.
\textit{iii) Joint 3D Semantic Occupancy and Flow Prediction.} This benchmark evaluates the joint prediction of semantics and motion on OpenOcc, following the standard protocol~\cite{tong2023scene,zhao20243D}. We report RayIoU for 16 semantic categories and mAVE\textsubscript{TP} for flow metrics in true-positive areas. Their weighted average is denoted as Occ Score. 
In particular, we observed that mAVE\textsubscript{TP} exhibits a strong dependence on semantic occupancy prediction, resulting in suboptimal convergence across epochs. Therefore, we additionally employ per-voxel mAVE as a supplementary metric.

\noindent\textbf{Implementation Details.}
Our default setup uses a ResNet-50 \cite{resnet} backbone with the input image size of $256 \times 704$. We also test with FlashInternImage-Tiny~\cite{xiong2024efficient} and scale up with Swin-Transformer Base \cite{liu2021Swin} at an image size of $512\times1408$. We follow established methods for depth estimation \cite{huang2022bevdet4d,li2023fb}, image/BEV augmentation \cite{huang2022bevdet4d}, and temporal fusion (16 frames) with CBGS strategies \cite{park2022time,li2023fb}. Our default 3D/2D models use voxel sizes of $200 \times 200 \times 16$ (32-dim features) and $200 \times 200 \times 1$ (80-dim features), respectively. All models are trained with the AdamW optimizer \cite{loshchilov2018fixing} with a learning rate of $2 \times 10^{-4}$ and a batch size of 16. Training lasts $12$ epochs for semantic occupancy and $18$ for joint occupancy semantic-flow prediction, with loss weights $\alpha=5, \beta=20$ (following Mask2Former~\cite{cheng2022masked}). \textbf{Notably}, the flow head and its supervision are used only for \textit{benchmark iii)}. Further details are in the supplementary material.

\subsection{Accuracy/Speed Trade-off Evaluation}

To assess ALOcc’s deployment capabilities, we analyze the speed/accuracy trade-off of its variants on \textit{benchmark ii)}. As shown in \cref{fig:trade-off}, we offer multiple model versions that achieve an excellent balance between accuracy and efficiency. Compared to ALOcc-3D, ALOcc-2D (-mini) employs spatial compression techniques similar to FlashOcc \cite{yu2023flashocc}, using height compression and 2D convolutions for volume encoding. ALOcc-2D-mini leverages monocular depth estimation \cite{li2022bevdepth} and reduced channel sizes, while other variants use stereo depth \cite{li2023bevstereo}. Results show that ALOcc-3D and ALOcc-2D surpass SOTAs with higher speeds, and ALOcc-2D-mini achieves real-time inference while maintaining a near-SOTA performance. Further details of each variant are provided in the \textbf{supplemental materials}.

\subsection{Comparison with SOTAs}
We compare our method with the current SOTAs on the Occ3D and OpenOcc datasets, with results summarized in \cref{tab:sota}, \cref{tab:sota-womask}, and \cref{tab:sem-flow}. ALOcc and ALOcc-Flow in the tables differ by the presence/absence of flow prediction. ALOcc-2D (default) uses 2D convolutions for height-compressed volume encoding with 80 channels, while ALOcc-3D (default) employs 3D convolutions for full 3D volume encoding with 32 channels. All results are reported without post-processing, such as test-time augmentation.

\noindent\textbf{3D Semantic Occupancy Prediction with Mask.}
\cref{tab:sota} compares ALOcc with SOTAs methods on Occ3D benchmark, focusing training on critical, camera-visible regions. Our method significantly outperforms existing approaches across various image input sizes and backbones. Notably, when provided with ground-truth depth in adaptive lifting, ALOcc surpasses even multi-modal fusion methods that rely on extra LiDAR or radar backbones, demonstrating the superiority of our approach.

\begin{table}[t]
  \centering
\resizebox{0.478\textwidth}{!}{
\setlength{\tabcolsep}{1.1mm}{

  \begin{tabular}{cl|c|c|c|c|ccc}
      \toprule
      Exp. & Condition &mIoU\textsuperscript{m}\textsubscript{D} &mIoU\textsuperscript{m} &mIoU & RayIoU& \multicolumn{3}{c}{RayIoU\textsubscript{1m, 2m, 4m}}   \\
      \midrule 
     0 & ALOcc-2D-40  & 38.5 & 44.5 & 36.8 &42.5 &36.6 &43.4 &47.7 \\
      1 & Exp.~0~\textit{w/o}~AL  & 37.5& 43.5& 36.1 &41.3& 35.2 & 42.1 &46.6 \\
      2 & Exp.~0~\textit{w/o}~SP &36.0 & 42.1 &33.8 &39.9 &34.5 &40.6 &44.7\\
      3 & Exp.~2~\textit{w/o}~AL  &34.9 &41.2 &33.3 &39.5 &34.0 &40.3 &44.2 \\
      
      \bottomrule
  \end{tabular}
  }
  }
  \vspace{-2mm}
  \caption{\textbf{Ablation study of semantic occupancy prediction on Occ3D.} \textbf{AL:} adaptive lifting; \textbf{SP:} semantic prototype-based occupancy head. ``\textit{w/o}" indicates the removal of a module.}
  \label{tab:occ_head_ablation}
  \vspace{-5.5mm}
\end{table}

\noindent\textbf{3D Semantic Occupancy Prediction without Mask.}
\cref{tab:sota-womask} showcases the results of 3D semantic occupancy prediction of training without the camera visible mask, where we outperform all prior methods on both mIoU and RayIoU.
This demonstrates the effectiveness of our method in capturing the complete scene context and understanding the spatial relationships between objects. 
Our method also exhibits superior performance in terms of RayIoU at different distances, further highlighting its robustness in capturing object details.
Notably, these performance gains are achieved with minimal computational overhead. Our scaled-down variant outperforms competing real-time models in inference speed while maintaining high accuracy, which is critical for on-device applications.

\noindent\textbf{Joint 3D Semantic Occupancy and Flow Prediction.}
\cref{tab:sem-flow} presents the results of 3D semantic occupancy and flow joint prediction on OpenOcc. Due to the limited existing approaches on this benchmark, we also include weakly supervised methods for comparison. Our method excels in Occ Score, mAVE, mAVE\textsubscript{TP}, and RayIoU across distances. It effectively captures semantic and motion information, demonstrating robustness in scene dynamics. ALOcc-Flow-2D outperforms ALOcc-Flow-3D in terms of mAVE and mAVE\textsubscript{TP}. This can be attributed to: 1) mAVE\textsubscript{TP} is computed on true positive semantic prediction areas, so higher semantic errors (ALOcc-Flow-2D) reduce the regions requiring this metric; 2) This benchmark evaluates the flow on X and Y axes, so height compression and added feature channels emphasize critical motion cues.

\subsection{Ablation Study}
Unless specified, we use ALOcc-2D with a BEV feature channel of 40 as our baseline for these experiments. 

\noindent\textbf{Components Analysis for Occ Head.}
\cref{tab:occ_head_ablation} presents an ablation study to evaluate the effectiveness of our core components in terms of semantic occupancy results. 
The full model (\textbf{Exp.~0}) demonstrates superior performance across all metrics. 
Removing the adaptive lifting component (\textbf{Exp.~1} and \textbf{Exp.~3}) results in decreased performance, particularly in mIoU\textsuperscript{m}\textsubscript{D} and mIoU, underscoring the efficacy of our adaptive and accurate 2D-to-3D view transformation. 
In \textbf{Exp.~2}, we substitute the semantic prototype-based occupancy head with FBOcc's occupancy head, leading to a more pronounced performance decline. 
 This highlights the crucial role of the SP component in enhancing the model's semantic occupancy prediction accuracy.

\begin{table}[t]
  \centering
\resizebox{0.478\textwidth}{!}{
\setlength{\tabcolsep}{1.1mm}{

  \begin{tabular}{cl|c|c|c|c|ccc}
      \toprule
      Exp.& Condition &Occ Score & mAVE& mAVE\textsubscript{TP} & RayIoU& \multicolumn{3}{c}{RayIoU\textsubscript{1m, 2m, 4m}}   \\
      \midrule 
      0 & ALOcc-2D-40  & - & -& -&42.4 &35.9 &43.3 &47.9 \\
      1 & Exp.~0~+~Flow  & 40.7 & 0.597& 0.508&39.7 &33.5 &40.5 &45.2 \\
      2 & Exp.~1~+~BC & 39.9& 0.565& 0.464 &38.3& 32.7 & 39.0 &43.3\\
      3 & Exp.~1~+~CV & 41.1 & 0.588 & 0.503 &40.2 & 33.8 &41.1 &45.6  \\
      4 & Exp.~2~+~CV & 41.1& 0.562 &0.451 &39.6&33.3 &40.4 &45.1\\
    5 & Exp.~4~+~CS &42.1  &0.537 & 0.427 &40.5& 34.3&41.3 &45.8\\
      \bottomrule
  \end{tabular}
  }
  }
 \vspace{-2mm}
  \caption{\textbf{Ablation study of 3D semantic occupancy and flow joint prediction results on OpenOcc.} \textbf{Flow}: flow head, \textbf{BC}: bin classification, \textbf{CV}: BEV cost volume, \textbf{CS}: channel scaling from 40 to 80. ``+'' indicates module addition.}
  \label{tab:flow_ablation}
  \vspace{-5mm}
\end{table}

\noindent\textbf{Components Analysis for Flow Head.}
As shown in \cref{tab:flow_ablation}, we conducted an ablation study to evaluate how different components affect flow prediction performance. In \textbf{Exp.~1}, we add a flow head implemented with a simple convolution. \textbf{Exp.~1} \vs \textbf{Exp.~0} show that adding the flow head harms the prediction of semantic occupancy. This is due to the motion flow containing different information from the semantic occupancy, which increases the burden of volume features. 
The results of \textbf{Exp.~2} \vs \textbf{Exp.~1} show that using {bin} classification significantly improves flow prediction performance. 
\textbf{Exp.~3} and \textbf{Exp.~4} demonstrate that the use of BEV cost volume alleviates pressure on scene representation, enhancing flow prediction while maintaining occupancy prediction performance. In \textbf{Exp.~5}, increasing BEV feature channels substantially improves overall performance, indicating that the information capacity of features is a bottleneck in joint learning of semantic occupancy and occupancy flow.

\vspace{-0.5ex}
\section{Conclusion}
In this paper, we explore the challenge of vision-based 3D semantic occupancy and flow prediction. We propose an occlusion-aware adaptive lifting method, complemented by depth denoising to enhance the adaptability and robustness of the 2D-to-3D view transformation process. To further improve semantic occupancy learning, we introduce a semantic prototype-based occupancy head that aligns 2D and 3D semantics, combined with hard sample mining techniques to mitigate the long-tail problem. Additionally, we present a BEV cost volume-based approach to facilitate occupancy flow learning, reducing the burden on features to represent semantics and motion simultaneously. Experiments conducted on the Occ3D and OpenOcc datasets demonstrate that our method outperforms current SOTA solutions. Benefiting from the lightweight nature of our approach, we provide multiple model versions: our highest-performing model is faster than other methods with comparable performance, while our fastest model achieves superior performance compared to methods with similar speed.

{ 
\small
    \bibliographystyle{ieeenat_fullname}
    \bibliography{main}
}

\clearpage
\setcounter{page}{1}
\appendix
\renewcommand\thefigure{A.\arabic{figure}} 
\renewcommand\theequation{A.\arabic{equation}} 
\renewcommand\thetable{A.\arabic{table}} 
\maketitlesupplementary

\begin{table*}[t]

  \setlength{\tabcolsep}{0.007\linewidth}
  \centering
  \resizebox{1.\textwidth}{!}{
  \begin{tabular}{l|c|c|c|c | c | c c c c c c c c c c c c c c c c c c}
      \toprule
     \multirow{-4.1}{*}{Method} & \multirow{-4.1}{*}{Input} & \multirow{-4.1}{*}{Backbone} & \multirow{-4.1}{*}{Image Size}
      & \rotatebox{90}{$\text{mIoU}_\text{D}^{\text{m}}$}
      & \rotatebox{90}{$\text{mIoU}^{\text{m}}$}
      & \rotatebox{90}{\textcolor{otherscolor}{$\blacksquare$} others} 
      & \rotatebox{90}{\textcolor{barriercolor}{$\blacksquare$} barrier} %
      & \rotatebox{90}{\textcolor{bicyclecolor}{$\blacksquare$} bicycle} %
      & \rotatebox{90}{\textcolor{buscolor}{$\blacksquare$} bus} %
      & \rotatebox{90}{\textcolor{carcolor}{$\blacksquare$} car} %
      & \rotatebox{90}{\textcolor{constructcolor}{$\blacksquare$} cons. veh.} %
      & \rotatebox{90}{\textcolor{motorcolor}{$\blacksquare$} motor.} %
      & \rotatebox{90}{\textcolor{pedestriancolor}{$\blacksquare$} pedes.} %
      & \rotatebox{90}{\textcolor{trafficcolor}{$\blacksquare$} tfc. cone} %
      & \rotatebox{90}{\textcolor{trailercolor}{$\blacksquare$} trailer} %
      & \rotatebox{90}{\textcolor{truckcolor}{$\blacksquare$} truck} %
      & \rotatebox{90}{\textcolor{driveablecolor}{$\blacksquare$} drv. surf.} %
      & \rotatebox{90}{\textcolor{otherflatcolor}{$\blacksquare$} other flat} %
      & \rotatebox{90}{\textcolor{sidewalkcolor}{$\blacksquare$} sidewalk} %
      & \rotatebox{90}{\textcolor{terraincolor}{$\blacksquare$} terrain} %
      & \rotatebox{90}{\textcolor{manmadecolor}{$\blacksquare$} manmade} %
      & \rotatebox{90}{\textcolor{vegetationcolor}{$\blacksquare$} vegetation} \\ %
      \midrule
       BEVDetOcc-SF~\cite{huang2022bevdet4d} &C& ResNet-50 & $256\times704$  & 34.4 & 41.9 &12.1&50.0&22.1&43.9&53.9&29.1&23.8&25.8&28.5&34.9&41.8&84.3&44.4&57.5&61.0&53.1&46.7 \\
        UniOCC~\cite{pan2023uniocc} &C& ResNet-50 & $256\times704$&- & 39.7 & - & - & - & - & - & - & - & - & - & - & - & - & - & - & - & - & - \\
        FB-Occ~\cite{li2023fbocc}&C& ResNet-50 & $256\times704$&34.2 &39.8&13.8&44.5&27.1&46.2&49.7&24.6&27.4&28.5&28.2&33.7&36.5&81.7&44.1&52.6&56.9&42.6&38.1 \\
        SurroundSDF~\cite{liu2024surroundsdf} &C& ResNet-50 & $256\times704$ &36.2 & 42.4&13.9&49.7&27.8&44.6&53.0&30.0&29.0&28.3&31.1&35.8&41.2&83.6&44.6&55.3&58.9&49.6&43.8 \\
        FlashOCC~\cite{yu2023flashocc} &C& ResNet-50 & $256\times704$  &24.7& 32.0&6.2&39.6&11.3&36.3&44.0&16.3&14.7&16.9&15.8&28.6&30.9&78.2&37.5&47.4&51.4&36.8&31.4 \\

         COTR~\cite{ma2023cotr}&C& ResNet-50 & $256\times704$&38.6 &44.5 & 13.3 & 52.1 & 32.0 & 46.0 & 55.6 & 32.6 & 32.8 & 30.4 & 34.1 & 37.7 & 41.8 & 84.5 & 46.2 & 57.6 & 60.7 & 52.0 & 46.3 \\
ViewFormer~\cite{li2024viewformer}  &C& ResNet-50 & $256\times704$ &35.0 & 41.9&12.9&50.1&28.0&44.6&52.9&22.4&29.6&28.0&29.3&35.2&39.4&84.7&\textbf{49.4}&57.4&59.7&47.4&40.6 \\
OPUS~\cite{wang2024opus} &C& ResNet-50 & $256\times704$  &33.3 &36.2 & 11.9 & 43.5 & 25.5 & 41.0 & 47.2 & 23.9 & 25.9 & 21.3 & 29.1 & 30.1 & 35.3 & 73.1 & 41.1 & 47.0 & 45.7 & 37.4 & 35.3 \\
\rowcolor{pink!10} \textbf{ALOcc-2D-mini} &C& ResNet-50 & $256\times704$ &35.4& 41.4 &14.2&48.6&28.7&44.8&52.8&24.7&29.2&29.0&32.0&34.6&39.6&82.4&46.9&54.8&57.7&44.7&39.3\\
            \rowcolor{pink!10} \textbf{ALOcc-2D} &C& ResNet-50 & $256\times704$ &38.7& 44.8 &\textbf{15.4}&52.2&\textbf{32.2}&46.2&55.4&28.2&\textbf{34.1}&\textbf{32.4}&\textbf{36.4}&38.0&42.8&84.2&48.8&57.4&60.0&52.9&45.6\\
           \rowcolor{pink!10} \textbf{ALOcc-3D}  &C& ResNet-50 & $256\times704$  &\textbf{39.3} &\textbf{45.5} &15.3&\textbf{52.5}&30.8&\textbf{47.2}&\textbf{55.9}&\textbf{32.7}&33.3&\textbf{32.4}&36.2&\textbf{38.9}&\textbf{43.7}&\textbf{84.9}&48.5&\textbf{58.8}&\textbf{61.9}&\textbf{53.5}&\textbf{47.3} \\

           \rowcolor{lightpink!8} \textbf{ALOcc-2D-mini}  &C& Intern-T & $256\times704$ &37.9& 43.7  &14.8&50.1&31.3&48.1&55.7&23.8&32.8&31.6&33.5&36.3&43.7&84.0&49.2&57.1&59.8&48.2&42.4\\
            \rowcolor{lightpink!8} \textbf{ALOcc-2D}  &C& Intern-T & $256\times704$ &40.7& 46.6 &16.3&53.3&35.0&48.3&57.6&28.7&35.3&34.6&38.1&40.0&46.5&85.2&50.5&59.1&61.8&54.3&47.1\\
           \rowcolor{lightpink!8} \textbf{ALOcc-3D}   &C& Intern-T & $256\times704$ &41.5&47.5 &
17.0&54.6&34.5&50.6&58.2&28.6&36.5&34.8&39.6&41.1&47.6&85.7&51.5&60.0&63.5&55.0&48.3\\
        \midrule
           HyDRa~\cite{wolters2024unleashing} &C+R& ResNet-50 & $256\times704$  &40.6 & 44.4 &15.1&51.1&32.7&52.3&56.3&29.4&35.9&35.1&33.7&39.1&44.1&80.4&45.1&52.0&55.3&52.1&44.4\\
           EFFOcc~\cite{shi2024effocc} &C+L& ResNet-50 & $256\times704$   &50.1 & 52.8 &12.1&\textbf{59.7}&33.4&\textbf{61.8}&\textbf{65.0}&35.5&\textbf{46.0}&\textbf{57.1}&41.0&47.9&\textbf{54.6}&82.8&44.0&56.4&60.2&\textbf{71.1}&\textbf{69.6}\\
           SDGOcc~\cite{duan2025sdgocc} &C+L& ResNet-50 & $256\times704$   &47.7 & 51.7 &13.2&57.8&24.3&60.3&64.3&36.2&39.4&52.4&35.8&\textbf{50.9}&53.7&84.6&47.5&58.0&61.6&70.7&67.7\\

           \rowcolor{pink!10} \textbf{ALOcc-2D-mini}  &C+D& ResNet-50 & $256\times704$ &46.2& 50.0 &15.7&54.6&36.6&55.7&60.6&34.8&41.0&44.9&39.3&44.5&51.1&83.6&48.5&57.3&60.2&62.7&58.2\\
            \rowcolor{pink!10} \textbf{ALOcc-2D}  &C+D& ResNet-50 & $256\times704$ &50.3& 53.5  &16.5&57.8&\textbf{41.6}&57.9&63.8&\textbf{37.6}&45.0&52.1&45.8&49.6&54.4&85.3&50.5&59.7&62.3&67.1&62.0\\
           \rowcolor{pink!10} \textbf{ALOcc-3D}   &C+D& ResNet-50 & $256\times704$ &\textbf{50.6} &\textbf{54.5} &\textbf{17.0}&59.0&40.9&58.3&64.4&37.2&45.9&52.7&\textbf{46.8}&50.5&54.5&\textbf{86.3}&\textbf{51.5}&\textbf{61.7}&\textbf{64.8}&69.1&65.1 \\
          \rowcolor{lightpink!8} \textbf{ALOcc-2D-mini}  &C+D& Intern-T & $256\times704$ &48.9& 52.1  &
17.4&56.5&39.2&60.4&62.8&34.7&45.2&45.9&41.1&48.6&54.5&85.3&50.3&59.5&62.3&63.2&58.8\\
            \rowcolor{lightpink!8} \textbf{ALOcc-2D}  &C+D& Intern-T & $256\times704$ &52.0& 54.9 &17.4&59.0&41.9&60.8&65.1&38.3&48.9&53.1&46.7&51.3&56.6&86.3&52.8&61.5&63.9&67.7&62.4\\
           \rowcolor{lightpink!8} \textbf{ALOcc-3D}   &C+D& Intern-T & $256\times704$ &52.4&55.6 &18.3&60.1&42.9&61.6&65.5&38.4&48.5&53.5&46.8&51.6&57.5&86.6&52.2&62.1&65.0&69.1&65.1\\
       \midrule
           \midrule
           BEVFormer~\cite{li2022bevformer} &C & ResNet-101 & $900\times1600$ &37.2 & 39.2  &5.0 & 44.9 & 26.2 & \textbf{59.7} & 55.1 & 27.9 & 29.1 & 34.3 & 29.6 & 29.1 & 50.5 & 44.4 & 22.4 & 21.5 & 19.5 & 39.3 & 31.1\\
           
           VoxFormer~\cite{li2023voxformer}  &C& ResNet-101 & $900\times1600$   &-& 40.7  & - & - & - & - & - & - & - & - & - & - & - & - & - & - & - & - & - \\
           SurroundOcc~\cite{wei2023surroundocc} &C & ResNet-50 & $900\times1600$   &31.2 & 37.2 &9.0 & 46.3 & 17.1 & 46.5 & 52.0 & 20.1 & 21.5 & 23.5 & 18.7 & 31.5 & 37.6 & 81.9 & 41.6 & 50.8 & 53.9 & 42.9 & 37.2\\
          
            FastOcc~\cite{hou2024fastocc} &C& ResNet-101 & $640\times1600$&34.5 &39.2  &2.1 & 43.5 & 28.0 & 44.8 & 52.2 & 23.0 & 29.1 & 29.7 & 27.0 & 30.8 & 38.4 & 82.0 & 41.9 & 51.9 & 53.7 & 41.0 & 35.5\\
            PanoOcc~\cite{wang2024panoocc} &C& ResNet-101 & $640\times1600$  &37.3 & 42.1  &11.7&50.5&29.6&49.4&55.5&23.3&33.3&30.6&31.0&34.4&42.6&83.3&44.2&54.4&56.0&45.9&40.4\\
            OSP~\cite{shi2024occupancy} &C& ResNet-101 & $900\times1600$  &37.0 & 41.2 &11.0 & 49.0 & 27.7 & 50.2 & 56.0 & 23.0 & 31.0 & 30.9 & 30.3 & 35.6 & 41.2 & 82.1 & 42.6 & 51.9 & 55.1 & 44.8 & 38.2\\
           BEVDetOcc~\cite{huang2022bevdet4d}  &C &Swin-Base & $512\times1408$   &36.9& 42.0 &12.2 & 49.6 & 25.1 & 52.0 & 54.5 & 27.9 & 28.0 & 28.9 & 27.2 & 36.4 & 42.2 & 82.3 & 43.3 & 54.6 & 57.9 & 48.6 & 43.6 \\
  
           COTR~\cite{ma2023cotr}  &C &Swin-Base & $512\times1408$  &41.3 &46.2  &14.9 & 53.3 & 35.2 & 50.8 & 57.3 & \bf35.4 & 34.1 & 33.5 & 37.1 & 39.0 & 45.0 & 84.5 & 48.7 & 57.6 & 61.1 & 51.6 & 46.7\\

           \rowcolor{pink!10} \textbf{ALOcc-2D} &C &Swin-Base & $512\times1408$ &44.5 &49.3 & 16.3 & 56.9 & 39.2 & 55.9 & 61.8 & 30.4 & 38.9 & 38.8 & 40.3 & 42.0 & 49.3 & 85.8 & 52.2 & 60.6 & 63.6 & 56.3 & 49.0 \\
           \rowcolor{pink!10} \textbf{ALOcc-3D}  &C &Swin-Base & $512\times1408$   &\textbf{46.1} & \textbf{50.6} & \textbf{17.0} & \textbf{58.3} & \textbf{39.7} & 56.6 & \textbf{63.2} & 33.2 & \textbf{41.3} & \textbf{40.3} & \textbf{40.8} & \textbf{43.7} & \textbf{51.0} & \textbf{87.0} & \textbf{52.7} & \textbf{62.0} & \textbf{65.1} & \textbf{57.7} & \textbf{50.9}\\
            \midrule
           OccFusion~\cite{ming2024occfusion} &C+L& ResNet-101 & $900\times1600$  & 45.3  &46.8&11.6&47.8&32.1&57.3&57.5&31.8&40.1&47.3&33.7&45.8&50.3&78.8&37.2&44.4&53.4&63.2&63.2\\
           BEVFusion~\cite{liu2022bevfusion} &C+L& Swin-Base & $512\times1408$  &48.7 & 54.0 &16.2&61.9&39.3&58.2&62.5&38.1&41.6&46.7&47.7&50.6&52.7&85.7&49.4&60.7&64.3&71.7&70.2\\
           EFFOcc~\cite{shi2024effocc} &C+L& Swin-Base & $512\times1408$  &50.7 & 54.1 &
15.7&61.0&36.2&62.2&66.4&38.7&43.9&52.1&42.4&50.3&56.1&84.9&48.0&58.6&62.0&71.3&69.5\\
FusionOcc~\cite{zhang2024fusionocc} &C+L& Swin-Base & $512\times1408$  &53.1 & 56.6 &17.1&62.6&43.1&63.8&66.2&37.9&49.7&53.7&49.8&53.1&57.5&86.2&49.8&61.6&65.1&73.5&\textbf{71.9}\\
\rowcolor{pink!10} \textbf{ALOcc-2D} &C+D&Swin-Base & $512\times1408$  &56.8 &58.7& 17.9&63.0&48.0&\textbf{66.1}&70.0&41.1&55.4&60.2&51.0&53.4&59.9&87.6&55.2&63.6&65.8&72.3&67.5\\
           \rowcolor{pink!10} \textbf{ALOcc-3D} &C+D &Swin-Base & $512\times1408$   &\textbf{57.8} & \textbf{60.0} & \textbf{18.7}&\textbf{64.6}&\textbf{50.5}&65.5&\textbf{70.9}&\textbf{42.1}&\textbf{56.2}&\textbf{61.5}&\textbf{52.6}&\textbf{54.4}&\textbf{61.5}&\textbf{88.3}&\textbf{55.3}&\textbf{64.8}&\textbf{67.9}&\textbf{74.2}&70.2 \\
  \bottomrule
  \end{tabular}
  }
\caption{\textbf{3D semantic occupancy prediction results \textit{\wrt} \textit{training \textit{with} camera visible mask} on Occ3D, showing per-class IoU, mIoU\textsubscript{D}\textsuperscript{m} and mIoU\textsuperscript{m}.} Input modalities include \textit{Camera (C)}, \textit{Radar (R)}, \textit{LIDAR (L)}, and \textit{Depth (D)}, where Depth represents a sparser signal mapped from LIDAR points. The best results among similar conditions (\ie, comparable image size, backbone, and input modalities) are \textbf{bolded}. ALOcc outperforms all competing methods in both mIoU\textsubscript{D}\textsuperscript{m} and mIoU\textsuperscript{m}, and also achieves the highest IoU scores for the majority of classes. Notably, even without a dedicated point cloud backbone, ALOcc achieves SOTA performance in multimodal scenarios.}
  \label{tab:occ_cls}
\end{table*}
   
\section{Metrics}

The Occ3D dataset \cite{tian2024occ3d} contains 18 categories, comprising 17 semantic categories: \textit{others}, \textit{barrier}, \textit{bicycle}, \textit{bus}, \textit{car}, \textit{construction vehicle}, \textit{motorcycle}, \textit{pedestrian}, \textit{traffic cone}, \textit{trailer}, \textit{truck}, \textit{driveable surface}, \textit{other flat}, \textit{sidewalk}, \textit{terrain}, \textit{manmade}, \textit{vegetation}, and an additional category representing non-occupied space, termed \textit{empty}. The OpenOcc dataset \cite{tong2023scene} contains all categories from Occ3D except for the \textit{others} category.
Our approach is evaluated across multiple metrics, including $\text{mIoU}^{\text{m}}_{\text{D}}$, $\text{mIoU}^m$, $\text{mIoU}$, \textit{RayIoU}, \textit{mAVE}, $\text{mAVE}_{\text{TP}}$, and \textit{Occ Score} \cite{liu2023fully,tong2023scene}. 

\textit{$\text{mIoU}^m$ and $\text{mIoU}$}: $\text{mIoU}^m$ and $\text{mIoU}$ represent the mean Intersection over Union (IoU) over all semantic categories (the superscript $\text{m}$ indicating whether the camera visibility mask was used during training):

\begin{equation*}
    \text{mIoU} = \frac{1}{|\mathcal{C}|} \sum_{c \in \mathcal{C}} \text{IoU}(c),
\end{equation*}
where $\mathcal{C}$ denotes the set of all semantic categories, and $\text{IoU}(c)$ represents the IoU for category $c$.

\textit{$\text{mIoU}^{\text{m}}_{\text{D}}$}: The $\text{mIoU}^{\text{m}}_{\text{D}}$ metric measures the mean IoU for the eight dynamic object categories (\ie, $\mathcal{C}_{\text{D}}$ = \{ \textit{bicycle}, \textit{bus}, \textit{car}, \textit{construction vehicle}, \textit{motorcycle}, \textit{pedestrian}, \textit{trailer}, \textit{truck} \}):
\begin{equation*}
    \text{mIoU}^{\text{m}}_{\text{D}} = \frac{1}{|\mathcal{C}_{\text{D}}|} \sum_{c \in \mathcal{C}_{\text{D}}} \text{IoU}(c).
\end{equation*}

\textit{RayIoU}: Ray-based mIoU~\cite{liu2023fully} calculates the mIoU using query rays instead of voxels, simulating LiDAR by projecting rays into the predicted 3D occupancy volume. A query ray is considered a true positive (TP) if both the predicted and ground-truth class labels match, and the L1 error between the predicted and ground-truth depth is within a certain threshold (\eg, 2m):
\begin{equation*}
    \text{RayIoU} = \frac{1}{|\mathcal{C}|} \sum_{c=1}^{|\mathcal{C}|} \frac{\text{TP}_c}{\text{TP}_c + \text{FP}_c + \text{FN}_c},
\end{equation*}
where $\text{TP}_c$ represents the true positives based on both semantic accuracy and depth error threshold, and $\text{FP}_c$ and $\text{FN}_c$ are the false positives and false negatives for class $c$, respectively. The final RayIoU is computed as the average of RayIoU values at thresholds of 1m, 2m, and 4m.

\textit{$\text{mAVE}_{\text{TP}}$}: The absolute velocity error (AVE) is defined for dynamic object categories $\mathcal{C}_{\text{D}}$. The $\text{mAVE}_{\text{TP}}$ is computed for the true positives of RayIoU when the depth error threshold is 2m, and represents the average velocity error for those voxels:
\begin{equation*}
    \text{mAVE}_{\text{TP}} = \frac{1}{|\mathcal{C}_{\text{D}}|} \sum_{c \in \mathcal{C}_{\text{D}}}\frac{1}{|\mathcal{V}_{\text{TP}}^{c}|} \sum_{v \in \mathcal{V}_{\text{TP}}^c} \text{AVE}(v),
\end{equation*}
where $\mathcal{V}_{\text{TP}}^c$ denotes the set of true positive voxels of class $c$.

\textit{$\text{mAVE}$}: The $\text{mAVE}$ is computed as the average velocity error across all voxels \textit{\wrt} dynamic object categories:
\begin{equation*}
    \text{mAVE} = \frac{1}{|\mathcal{C}_{\text{D}}|} \sum_{c \in \mathcal{C}_{\text{D}}}\frac{1}{|\mathcal{V}^{c}|} \sum_{v \in \mathcal{V}^c} \text{AVE}(v).
\end{equation*}

\textit{Occ Score}: The occupancy score is a comprehensive metric for joint evaluation of semantic occupancy and motion flow, defined as a weighted sum of RayIoU and $\text{mAVE}_\text{TP}$. The Occ Score is given by:
\begin{equation*}
    \text{OccScore} = \text{RayIoU} \times 0.9 + \text{max}(1 - \text{mAVE}_\text{TP}, 0.0) \times 0.1.
\end{equation*}

\begin{table*}[!ht]

  \setlength{\tabcolsep}{0.007\linewidth}
  \centering
  \resizebox{1.\textwidth}{!}{
  \begin{tabular}{l|c|c|c | c | c | c c c c c c c c c c c c c c c c}
      \toprule
      \multirow{-4.1}{*}{Method}
      & \multirow{-4.1}{*}{Backbone}
      & \multirow{-4.1}{*}{Image Size}
      
      & \rotatebox{90}{$\text{mIoU}_\text{D}$}
      & \rotatebox{90}{$\text{mIoU}$}
      
      & \rotatebox{90}{IoU}
      & \rotatebox{90}{\textcolor{barriercolor}{$\blacksquare$} barrier} %
      & \rotatebox{90}{\textcolor{bicyclecolor}{$\blacksquare$} bicycle} %
      & \rotatebox{90}{\textcolor{buscolor}{$\blacksquare$} bus} %
      & \rotatebox{90}{\textcolor{carcolor}{$\blacksquare$} car} %
      & \rotatebox{90}{\textcolor{constructcolor}{$\blacksquare$} cons. veh.} %
      & \rotatebox{90}{\textcolor{motorcolor}{$\blacksquare$} motor.} %
      & \rotatebox{90}{\textcolor{pedestriancolor}{$\blacksquare$} pedes.} %
      & \rotatebox{90}{\textcolor{trafficcolor}{$\blacksquare$} tfc. cone} %
      & \rotatebox{90}{\textcolor{trailercolor}{$\blacksquare$} trailer} %
      & \rotatebox{90}{\textcolor{truckcolor}{$\blacksquare$} truck} %
      & \rotatebox{90}{\textcolor{driveablecolor}{$\blacksquare$} drv. surf.} %
      & \rotatebox{90}{\textcolor{otherflatcolor}{$\blacksquare$} other flat} %
      & \rotatebox{90}{\textcolor{sidewalkcolor}{$\blacksquare$} sidewalk} %
      & \rotatebox{90}{\textcolor{terraincolor}{$\blacksquare$} terrain} %
      & \rotatebox{90}{\textcolor{manmadecolor}{$\blacksquare$} manmade} %
      & \rotatebox{90}{\textcolor{vegetationcolor}{$\blacksquare$} vegetation} \\ %
      \midrule
     
BEVFormer~\cite{li2022bevformer}& ResNet-101 & $900\times1600$  &14.2  & 16.8& 30.5 &14.2&6.6&23.5&28.3&8.7&10.8&6.6&4.1&11.2&17.8&37.3&18.0&22.9&22.2&13.8&22.2\\

            TPVFormer~\cite{huang2023tri} & ResNet-101 & $900\times1600$   &14.0& 17.1 & 30.9 &16.0&5.3&23.9&27.3&9.8&8.7&7.1&5.2&11.0&19.2&38.9&21.3&24.3&23.2&11.7&20.8

 \\
            SurroundOcc~\cite{wei2023surroundocc}& ResNet-101 & $900\times1600$   &18.4& 20.3 & 31.5 &20.6&11.7&28.1&30.9&10.7&15.1&14.1&12.1&14.4&22.3&37.3&23.7&24.5&22.8&14.9&21.9  \\

            GaussianFormer~\cite{huang2024gaussianformer}   &ResNet-101 & $900\times1600$& 19.1 &17.3 & 29.8 &19.5&11.3&26.1&29.8&10.5&13.8&12.6&8.7&12.7&21.6&39.6&23.3&24.5&23.0&9.6&19.1  \\
            GaussianFormer2~\cite{huang2025gaussianformer}& ResNet-101 & $900\times1600$    &18.8& 20.8 & 31.7 &21.4&13.4&28.5&30.8&10.9&15.8&13.6&10.5&14.0&22.9&40.6&24.4&26.1&24.3&13.8&22.0  \\
            
            BEVDetOcc~\cite{huang2022bevdet4d} & ResNet-50 & $900\times1600$ &14.1& 17.5 & 29.2 
&18.1&2.1&25.5&29.5&11.6&9.5&7.0&4.4&7.3&20.1&40.4&21.3&26.3&23.8&11.5&21.6 \\
     
           \rowcolor{pink!10} \textbf{ALOcc-2D-mini*}& ResNet-50 & $900\times1600$    &19.5& 21.5 &31.5  &21.8&15.7&27.3&30.7&12.7&17.4&15.7&14.0&13.9&22.4&40.0&24.7&26.3&24.4&14.4&22.3\\
          \rowcolor{pink!10} \textbf{ALOcc-2D*}& ResNet-50 & $900\times1600$    &21.5& 23.7 &34.5  
&23.5&17.2&28.0&33.0&17.0&19.2&17.1&16.1&15.0&25.1&41.9&26.8&28.3&26.5&18.5&26.5\\
           \rowcolor{pink!10} \textbf{ALOcc*} & ResNet-50 & $900\times1600$  &21.7 &24.0& 34.7 &23.8&17.4&28.0&32.9&17.0&20.2&17.2&\textbf{16.9}&15.4&25.5&41.8&26.7&28.3&27.0&18.6&27.0 \\

          \rowcolor{pink!10} \textbf{ALOcc-2D-mini*}&Intern-T & $900\times1600$    &21.0&23.3&33.6&23.4&16.5&29.1&33.0&12.3&20.0&17.4&15.6&15.1&24.5&41.3&26.3&28.2&26.6&17.2&25.4\\
          \rowcolor{pink!10} \textbf{ALOcc-2D*}& Intern-T & $900\times1600$    &22.6&24.9&35.8&\textbf{25.0}&18.7&29.7&34.1&16.1&\textbf{22.0}&17.8&16.6&15.8&\textbf{26.3}&42.5&27.9&29.4&28.0&20.1&28.4\\
           \rowcolor{pink!10} \textbf{ALOcc*} & Intern-T & $900\times1600$  &\textbf{22.8}&\textbf{25.1}&\textbf{36.1}&24.8&\textbf{19.0}&\textbf{30.2}&\textbf{34.3}&\textbf{17.5}&21.4&\textbf{18.2}&\textbf{16.9}&\textbf{15.9}&\textbf{26.3}&\textbf{42.9}&\textbf{28.0}&\textbf{29.6}&\textbf{28.2}&\textbf{20.2}&\textbf{28.8}\\

  \bottomrule
  \end{tabular}
  }
\caption{\textbf{3D semantic occupancy prediction results on SurorundOcc, reporting mIoU, mIoU\textsubscript{D}, and per-class IoU.} The metrics are computed over all voxels, a notable difference from the Occ3D benchmark. For a fair comparison against existing methods, we report results without voxel-level temporal fusion (denoted by *). Our approach sets a new SOTA on this benchmark, surpassing all competing methods on every metric.}
\vspace{1ex}
  \label{tab:occ_cls_surroundocc}
\end{table*}

\begin{table*}[t]
  \centering

  \renewcommand{\arraystretch}{1.1}
  \resizebox{1.\textwidth}{!}{
  \begin{tabular}{l|c|*{15}{c}}
   \toprule
    Method & \rotatebox{90}{ mIoU} & \rotatebox{90}{ Go} & \rotatebox{90}{ Vehicle} & \rotatebox{90}{ Pedestrian} & \rotatebox{90}{ Sign} & \rotatebox{90}{ Bicyclist} & \rotatebox{90}{ Traffic Light} & \rotatebox{90}{ Pole} & \rotatebox{90}{ Cons. Cone} & \rotatebox{90}{ Bicycle} & \rotatebox{90}{ Building} & \rotatebox{90}{ Vegetation} & \rotatebox{90}{ Tree Trunk} & \rotatebox{90}{ Road} & \rotatebox{90}{ Walkable} 
    \\
    \midrule
    BEVFormer-w/o TSA
    & 23.87 & \textbf{7.50} & 34.54 & 21.07 & 9.69 & \textbf{20.96} & 11.48 & 11.48 & 14.06 & 14.51 & 23.14 & 21.82 & 8.57 & 78.45 & 56.89 \\
    BEVFormer~\cite{li2022bevformer}  & 24.58 & 7.18 & 36.06 & 21.00 & 9.76 & 20.23 & 12.61 & 14.52 & 14.70 & 16.06 & 23.98 & 22.50 & 9.39 & 79.11 & 57.04  \\
    SOLOFusion~\cite{park2022time} & 24.73 & 4.97 & 32.45 & 18.28 & 10.33 & 17.14 & 8.07 & 17.83 & 16.23 & 19.3 & 31.49 & 28.98 & 16.93 & 70.95 & 53.28 \\
    BEVFormer-WrapConcat & 25.07 & 6.2 & 36.17 & 20.95 & 9.56 & 20.58 & 12.82 & 16.24 & 14.31 & 16.78 & 25.14 & 23.56 & 12.81 & 79.04 & 56.83  \\
    CVT-Occ~\cite{ye2024cvt} & 27.37 & 7.44 & \textbf{41.0} & 23.93 & 11.92 & 20.81 & 12.07 & 18.03 & 16.88 & \textbf{21.37} & 29.4 & 27.42 & 14.67 & \textbf{79.12} & 59.09 \\
    
    \rowcolor{pink!10} \textbf{ALOcc-3D} & \textbf{30.03} & 6.51 & 39.61 & \textbf{24.14} & \textbf{20.84} & 20.56 & \textbf{20.56} & \textbf{24.28} & \textbf{17.95} & 12.22 &  \textbf{35.67} &  \textbf{37.25} & \textbf{22.45} & 78.42& \textbf{59.91} \\
    \bottomrule
  \end{tabular}
  }
  \caption{\textbf{3D semantic occupancy prediction results on Occ3D-Waymo.} All methods use image input. We use an input image size of $640\times960$ and a backbone of ResNet-50 for comparison with other methods, with settings for other approaches detailed in CVT-Occ \cite{ye2024cvt}.}
  \label{tab:waymo_results}
\end{table*}

\begin{figure*}[t]
    \centering
    \setlength{\abovecaptionskip}{0pt}
    \includegraphics[width=1\linewidth]{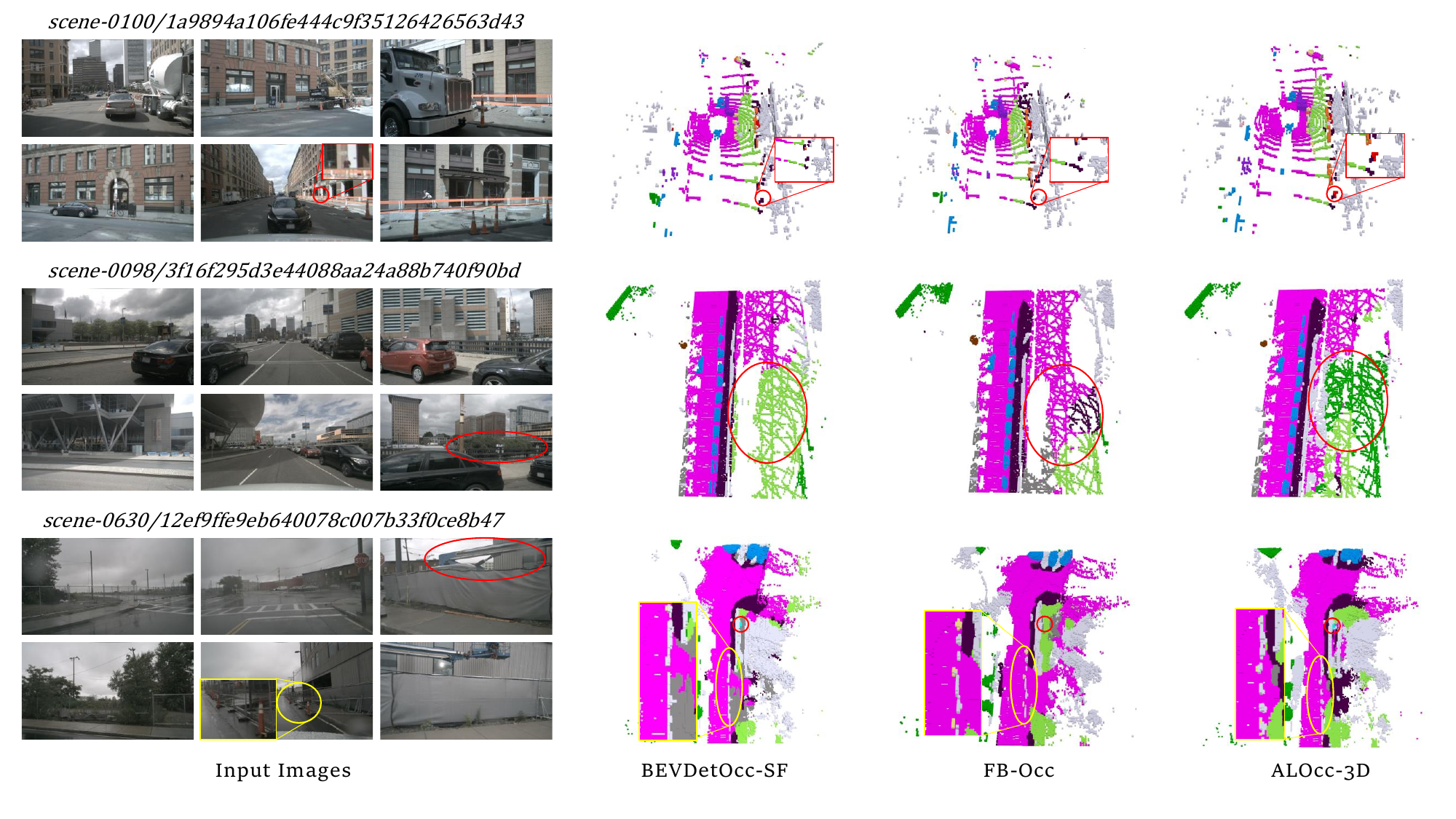}
    \caption{\textbf{Qualitative results on Occ3D.} From left to right, the columns represent the input images, visualization results of BEVDetOcc-SF, FB-Occ, and our ALOcc-3D. Key differences are highlighted in red or orange. The corresponding categories for different colors can be found in \cref{tab:occ_cls}.}
    \vspace{1ex}
    \label{fig:occ}
\end{figure*}

\section{Additional Experimental Results}
\subsection{Per-Class IoU}
\cref{tab:occ_cls} presents the per-class 3D occupancy prediction results \textit{\wrt} \textit{training with camera visible mask} on Occ3D. ALOcc achieves the best performance in most classes. Notably, ALOcc demonstrates significant improvements in rare but traffic-critical categories such as \textit{pedestrian} and \textit{truck}, highlighting the practical applicability of our method in real-world scenarios.
\subsection{Results on Surroundocc Benchmark}
We additionally evaluate our method on the SurroundOcc benchmark~\cite{wei2023surroundocc} to facilitate a broader comparison with recent work. Although also built on the nuScenes dataset, SurroundOcc differs from Occ3D by omitting the ``others'' semantic class and providing ground truth annotations of slightly lower quality. Despite this, its adoption in many recent studies~\cite{wei2023surroundocc,huang2024gaussianformer,huang2025gaussianformer} makes it a relevant benchmark. A key characteristic of its evaluation protocol is that, unlike Occ3D, metrics are computed over all voxels. To ensure a fair comparison with prior work that seldom incorporates temporal fusion, we report our results without this component, marking them with an asterisk (*). As shown in \cref{tab:occ_cls_surroundocc}, ALOcc achieves SOTA performance, outperforming all other competitors across all reported metrics.

\subsection{Results on Occ3D-Waymo}

To validate the generalization capability of our method, we conduct experiments on the large-scale Occ3D-Waymo dataset~\cite{tian2024occ3d}. This dataset is significantly more extensive than nuScenes, containing $5.6\times$ more training frames. This dataset presents a significant challenge that few methods have reported on at full scale. For this evaluation, we use a ResNet-50 backbone and follow CVT-Occ~\cite{ye2024cvt} in setting the input image resolution to $640\times960$, while all other hyperparameters remain consistent with our nuScenes experiments. As presented in~\cref{tab:waymo_results}, our method surpasses all competing approaches, underscoring its robust generalization to a different and more challenging scenario.

\subsection{Visualization}
As shown in~\cref{fig:occ}, we provide qualitative visualizations to compare our 3D semantic occupancy predictions against other methods. The results highlight several key advantages of our approach. In the first row, our model demonstrates superior sensitivity by successfully identifying pedestrians, which competing methods overlook. The second row showcases its ability to reconstruct large, complex structures, accurately capturing a dense cluster of trees that are largely omitted by the others. Finally, the third row underscores our method's robustness across different object scales: it precisely predicts both a small traffic cone, which BEVDetOcc-SF fails to render correctly, and a large construction vehicle missed entirely by FB-Occ. Collectively, these visualizations demonstrate that ALOcc produces more complete and semantically accurate scene representations.

\subsection{Experiments on Model Architecture}

\paragraph{Regarding Adaptive Lifting.}
As shown in \cref{tab:al}, we conducted experiments on the benchmark of \textit{training without mask} to investigate the impact of the adaptive lifting module. The results indicate that converting depth weights to occlusion weights significantly improves the prediction of dynamic objects, while depth denoising further enhances overall performance. Furthermore, we set the value of $m$ to 3 for managing inter-object occlusion, based on the experimental results in ~\cref{tab:miou_k}.

As illustrated in~\cref{tab:dd}, we analyze the effect of different epochs for depth denoising on the performance. The parameter $E$ in Eq.~(\textcolor{green}{5}) was set as the number of epochs multiplied by the number of iterations per epoch. Our default model uses an epoch number of 6.

\begin{table}[t]
  \centering
  \resizebox{0.35\textwidth}{!}{
      \begin{tabular}{cl|c|c}
        \toprule
       Exp. & Condition & $\text{mIoU}^{m}_{D}$ & $\text{mIoU}^{m}$ \\
        \midrule
        0 & ALOcc-2D-40 (\textit{w/o} SP) & 36.0 & 42.1 \\
        1 & Exp.~0 \textit{w/o} DD  & 35.8 & 41.8 \\
        2 & Exp.~1 \textit{w/o} D2IO  & 35.7 & 41.6 \\
        3 & Exp.~1 \textit{w/o} D2O  & 34.9 & 41.2 \\
        \bottomrule
      \end{tabular}
    }
  \caption{\textbf{Ablation study on adaptive lifting.} \textbf{SP} denotes the Semantic Prototype-based occupancy head. \textbf{DD} represents depth denoising. \textbf{D2O} indicates converting depth weights to occlusion weights. \textbf{D2IO} refers to converting depth weights to inter-object occlusion.}
  \label{tab:al}
\end{table}

\begin{table}[h]
\centering
\resizebox{0.22\textwidth}{!}{
\begin{tabular}{c|c|c}
        \toprule
        Num. & $\text{mIoU}^{m}_{D}$ & $\text{mIoU}^{m}$ \\
        \midrule
        0 &35.7& 41.6 \\
        1 &35.9& 41.9 \\
3 &36.0& 42.1 \\
5 &35.7& 41.8 \\
        \bottomrule
      \end{tabular}
      }
\caption{\textbf{Effects of different $m$ values (regarding inter-object occlusion).}}
\label{tab:miou_k}
\end{table}

\begin{table}[t]
  \centering
  \resizebox{0.22\textwidth}{!}{
      \begin{tabular}{c|c|c}
        \toprule
        Num. & $\text{mIoU}^{m}_{D}$ & $\text{mIoU}^{m}$ \\
        \midrule
        2 & 38.1 & 44.2 \\
        4 & 38.3 & 44.4 \\
        6 & 38.5 & 44.5 \\
        8 & 38.5 & 44.5 \\
        \bottomrule
      \end{tabular}
    }
  \vspace{-0ex}
  \caption{\textbf{Effects of depth denoising \textit{w.r.t.} the \textit{number of epochs} of conducting depth denoising.}}
  \label{tab:dd}
  \vspace{-0mm}
\end{table}

\begin{table}[t]
  \centering
  \resizebox{0.22\textwidth}{!}{
      \begin{tabular}{c|c|c}
        \toprule
         Num & $\text{mIoU}^{m}_{D}$ & $\text{mIoU}^{m}$ \\
        \midrule
        1$\times$ & 37.8& 43.9   \\
        2$\times$ & 38.5 & 44.5 \\
        4$\times$ & 38.0 & 44.2\\
        \bottomrule
      \end{tabular}
  }
  \vspace{-0ex}
  \caption{\textbf{Effects of point \textbf{sampling number} $K$.} The notations $1\times$, $2\times$, and $4\times$ represent a multiplication factor of 12544.}
  \label{tab:num}
  \vspace{-0mm}
\end{table}

\paragraph{Effect of Point Sampling Density in Supervision.}

We conduct an ablation study in~\cref{tab:num} to analyze the impact of the number of sampled points, $K$, used during occupancy supervision. Using the 12544 points sampled by default in Mask2Former~\cite{cheng2022masked} as a baseline ($1\times$), we experiment with sampling factors of $2\times$ and $4\times$. Based on the results, we adopt a sampling density of $2\times$ (25088 points) as the default setting for our models, as it provides a favorable balance of performance and efficiency.

\section{Additional Experimental Details}

\subsection{Model Details}
We primarily provide three models: ALOcc-3D, ALOcc-2D, and ALOcc-2D-mini. The main difference between ALOcc-3D and ALOcc-2D lies in the processing before the volume encoder. ALOcc-2D compresses the height of the volume features into the channel dimension before passing it to the volume encoder, utilizing 2D convolution for feature encoding \cite{yu2023flashocc}. Before the prediction head, a simple convolution layer is used to recover the height dimension from the channel dimension, thus avoiding the high computational cost of 3D convolution. For the ResNet-50 and the Intern-T baselines, during feature encoding, ALOcc-3D has feature dimensions of $200 \times 200 \times 16 \times 32$, while ALOcc-2D has feature dimensions of $200 \times 200 \times 80$. ALOcc-2D-mini further simplifies the depth prediction module by switching from stereo depth estimation to single-view depth estimation and using a smaller channel dimension of 40. For the Swin-Base baseline, ALOcc-3D uses a channel dimension of 64, while ALOcc-2D uses a channel dimension of 160.

\subsection{Training Details}
Our models are initialized using publicly available checkpoints: we adopt the BEVDet~\cite{huang2022bevdet4d} checkpoint for ResNet-50 backbones and the GeoMIM~\cite{Liu_2023_ICCV} checkpoint for Swin-Base backbones. For additional experiments on the Intern-T backbone, we pre-train it ourselves with BEVDet. The models in Tab.~\textcolor{green}{1} and Tab.~\textcolor{green}{2} differ in whether a camera-visible mask was used during training. Models in Tab.~\textcolor{green}{1} use this mask, constraining their objective to mapping observed image content into 3D space. In contrast, models in Tab.~\textcolor{green}{2} are trained without this constraint, tasking them with the more challenging goal of inferring both visible and occluded content. Additionally, our models in Tab.~\textcolor{green}{3} use a ray-visible mask \cite{chen2024adaocc} during training, which was generated with ray queries.

\section{Data Source}
To facilitate data traceability, we additionally document the data sources for the compared methods presented in Tab.~\textcolor{green}{1}, Tab.~\textcolor{green}{2}, and Tab.~\textcolor{green}{3}.
The results of BEVFormer and RenderOcc (Tab.~\textcolor{green}{2}) were cited from SparseOcc~\cite{liu2023fully}, while the results of SurroundOcc were cited from FastOcc \cite{hou2024fastocc}. The results of VoxFormer were cited from COTR~\cite{ma2023cotr}. The results of OccFormer, TPVFormer, and CTF-Occ were cited from Occ3D~\cite{tian2024occ3d}. The results of BEVFusion were cited from FusionOcc~\cite{zhang2024fusionocc}. The results of OccNet, OccNerf, and RenderOcc (Tab.~\textcolor{green}{3}) were cited from LetOccFlow~\cite{liu2024letoccflow}. The results of UniOCC, SurroundSDF, COTR, HyDRa, FastOcc, PanoOcc, EFFOcc, SDGOcc, OccFusion, FusionOcc, SparseOcc, LetOccFlow, CascadeFlow, and F-Occ were cited from their original papers.
The results of FlashOcc, Panoptic-FlashOCC, ViewFormer, OPUS, OSP, and BEVDetOcc were evaluated using the official checkpoints~\cite{yu2023flashocc, yu2024panoptic,li2024viewformer,wang2024opus,shi2024occupancy,huang2022bevdet4d}. We reproduced the results for FBOcc using the official code. BEVDetOcc-SF was implemented by ourselves. We extended it using SoloFusion~\cite{park2022time} to create a long history (16-frame) fusion version. All FPS were measured by ourselves using a single RTX 4090 GPU.

\section{Definition of the Occluded Length}
We give a mathematical formulation of the occluded length.
\begin{definition}[Occluded Length.]
    \label{def:hl}
    Let $(u, v, d) \in \mathbb{R}^3$ represent the coordinates of the surface point in the camera frustum space, with $(u,v)$ denoting the coordinate of the corresponding pixel and $d$ representing depth. The Occluded length $l$ at this point is defined as the maximal extension within the object's range from the camera's perspective. It satisfies the following conditions:
    \begin{enumerate}
        \item $\forall \lambda \in (0, 1], \mathrm{CLASS}(u, v, d) = \mathrm{CLASS}(u, v, d + \lambda l)$,
        \item There exists a unique $l \in \mathbb{R}_+$ such that:
        
        $\lim_{\epsilon \to 0^+} \mathrm{CLASS}(u, v, d + l) \neq \mathrm{CLASS}(u, v, d + l + \epsilon)\} $,
    \end{enumerate}
    where $\mathrm{CLASS} : \mathbb{R}^3 \rightarrow \mathcal{C}$ maps frustum space coordinates to the class space $\mathcal{C}$.
    \end{definition}


\end{document}